\documentclass[a4paper,english]{article}
\usepackage[T1]{fontenc}
\usepackage[latin1]{inputenc}
\usepackage{color}
\usepackage{float}
\usepackage{graphicx}
\usepackage{amssymb}

\floatstyle{ruled}
\newfloat{algorithm}{tbp}{loa}
\floatname{algorithm}{Algorithm}

 \usepackage{jmlr2e-lyx}
 \usepackage{amsmath}
 \usepackage{amsfonts}

\usepackage{amsfonts}
\usepackage{amssymb}
\usepackage[mathscr]{eucal}
\usepackage[round]{natbib}

\usepackage{babel}

\begin{document}
\jmlrheading{}{}{}{02/01}{}{D.~Di~Castro and R. Meir 2009}

\ShortHeadings{A Convergent Online Single Time Scale Actor Critic
Algorithm}{Di Castro and Meir}

\title{A Convergent Online Single Time Scale Actor Critic Algorithm}

\author{\name Dotan Di Castro \email dot@tx.technion.ac.il \name
\AND \name Ron Meir \email rmeir@ee.technion.ac.il
\AND \name \addr Department of Electrical Engineering, Technion, Haifa 32000,
Israel}

\editor{}
\maketitle
\begin{abstract}
Actor-Critic based approaches were among the first to address reinforcement
learning in a general setting. Recently, these algorithms have gained
renewed interest due to their generality, good convergence properties,
and possible biological relevance. In this paper, we introduce an
online temporal difference based actor-critic algorithm which is proved
to converge to a neighborhood of a local maximum of the average reward.
Linear function approximation is used by the critic in order estimate
the value function, and the temporal difference signal, which is passed
from the critic to the actor. The main distinguishing feature of the
present convergence proof is that both the actor and the critic operate
on a similar time scale, while in most current convergence proofs
they are required to have very different time scales in order to converge.
Moreover, the same temporal difference signal is used to update the
parameters of both the actor and the critic. A limitation of the proposed
approach, compared to results available for two time scale convergence,
is that convergence is guaranteed only to a neighborhood of an optimal
value, rather to an optimal value itself. The single time scale and
identical temporal difference signal used by the actor and the critic,
may provide a step towards constructing more biologically realistic
models of reinforcement learning in the brain.
\end{abstract}
\global\long\def\US{\mathcal{U}}

\global\long\def\XS{\mathcal{X}}

\global\long\def\br{\bar{r}}

\global\long\def\E{\textrm{E}}

\global\long\def\R{\mathbb{R}}

\global\long\def\RK{\textrm{\mathbb{R}}^{K}}

\global\long\def\PS{\mathcal{P}}

\global\long\def\he{\tilde{h}}

\global\long\def\de{\tilde{d}}

\global\long\def\ee{\tilde{\eta}}

\global\long\def\EtaE{\tilde{\eta}}

\global\long\def\nt{\nabla_{\theta}}

\global\long\def\nw{\nabla_{w}}

\global\long\def\one{\mathbf{\boldsymbol{1}}}

\global\long\def\F{\mathcal{F}}

\global\long\def\FF{\bar{\mathcal{F}}}

\global\long\def\M{\mathcal{M}}

\global\long\def\Bgradeta{B_{\nabla\eta}}

\section{Introduction\label{sec:Introduction}}

In Reinforcement Learning (RL) an agent attempts to improve its performance
over time at a given task, based on continual interaction with the
(usually unknown) environment (\cite{BerTsi96,SutBar98}). Formally,
it is the problem of mapping situations to actions in order to maximize
a given average reward signal. The interaction between the agent and
the environment is modeled mathematically as a Markov Decision Process
(MDP). Approaches based on a direct interaction with the environment,
are referred to as \emph{simulation based algorithms}, and will form
the major focus of this paper.

A well known subclass of RL approaches consists of the so called actor-critic
(AC) algorithms (e.g., \cite{SutBar98}), where the agent is divided
into two components, an actor and a critic. The critic functions as
a value estimator, whereas the actor attempts to select actions based
on the value estimated by the critic. These two components solve their
own problems separately but interactively. Many methods for solving
the critic's value estimation problem, for a \emph{fixed} policy,
have been proposed, but, arguably, the most widely used is \emph{temporal
difference} (TD) learning. TD learning was demonstrated to accelerate
convergence by trading bias for variance effectively \cite{SinDay98},
and is often used as a component of AC algorithms.

In general, policy selection may be randomized. When facing problems
with a large number of states or actions (or even continuous state-action
problems), effective policy selection may suffer from several problems,
such as slow convergence rate or an inefficient representation of
the policy. A possible approach to policy learning is the so-called
\emph{policy gradient method} (\cite{BaxBar01,Cao07,CaoChen1997,KonTsi03,MarTsi98}).
Instead of maintaining a separate estimate for the value for each
state (or state-action pair), the agent maintains a parametrized policy
function. The policy function is taken to be a differentiable function
of a parameter vector and of the state. Given the performance measure,
depending on the agent's policy parameters, these parameters are updated
using a sampling-based estimate of the gradient of the average reward.
While such approaches can be proved to converge under certain conditions
(e.g., \cite{BaxBar01}), they often lead to slow convergence, due
to very high variance. A more general approach based on sensitivity
analysis, which includes policy gradient methods as well as non-parametric
average reward functions, has been discussed in depth in the recent
manuscript by \cite{Cao07}.

Several AC algorithms with associated convergence proofs have been
proposed recently (a short review is given in section \ref{sec:previous_AC}).
As far as we are aware, all the convergence results for these algorithms
are based on two time scales, specifically, the actor is assumed to
update its internal parameters on a much slower time scale than the
one used by the critic. The intuitive reason for this time scale separation
is clear, since the actor improves its policy based on the critic's
estimates. It can be expected that rapid change of the policy parameters
may not allow the critic to effectively evaluate the value function,
which may lead to instability when used by the actor in order to re-update
its parameters.

The objective of this paper is to propose an online AC algorithm and
establish its convergence under conditions which do \emph{not} require
the separation into two time scales. There is clear theoretical motivation
for such an approach,\textcolor{magenta}{ }as it can potentially lead
to faster convergence rates, although this is not a an issue we stress
in this work. In fact, our motivation for the current direction was
based on the possible relevance of AC algorithms in a biological context
(e.g, \cite{NivDawDayan2006}), where it would be difficult to justify
two very different time scales operating within the same anatomical
structure. We refer the reader to \cite{DICasVolMei08} for some preliminary
ideas and references related to these issues. Given the weaker conditions
assumed on the time scales, our convergence result is, not surprisingly,
somewhat weaker than that provided recently in (e.g., \cite{BhaSutGhaLee08a,BhaSutGhaLee08b}),
as we are not ensured to converge to a local optimum, but only to
a neighborhood of such an optimum. Nevertheless, it is shown that
the neighborhood size can be algorithmically controlled. Further comparative
discussion can be found in section \ref{sec:previous}.

This paper is organized as follows. In section \ref{sec:previous}
we briefly recapitulate current AC algorithms for which convergence
proofs are available. In section \ref{sec:ProblemSetup}, we formally
introduce the problem setup. We begin section \ref{sec:MainAlgorithm}
by relating the TD signal to the gradient of the average reward, and
then move on to motivate and derive the main AC algorithm, concluding
the section with a convergence proof. A comparative discussion of
the main features of our approach is presented in section \ref{sec:comparison},
followed by some simulation results in section \ref{sec:simulations}.
Finally, in section \ref{sec:Discussion}, we discuss the results
and point out possible future work. In order to facilitate the readability
of the paper, we have relegated all technical proofs to appendices.

\section{Previous Work\label{sec:previous}}

In this section we briefly review some previous work in RL which bears
direct relevance to our work. While many AC algorithms have been introduced
over the years, we focus only on those for which a convergence proof
is available, since the main focus of this work is on convergence
issues, rather than on establishing the most practically effective
algorithms (see, for example, \cite{PetSch08}, for promising applications
of AC algorithms in a robotic setting).

\subsection{Direct policy gradient algorithms}

Direct policy gradient algorithms, employing agents which consist
of an actor only, typically estimate a noisy gradient of the average
reward, and are relatively close in their characteristics to AC algorithms.
The main difference from the latter is that the agent does not maintain
a separate value estimator for each state, but rather interacts with
the environment directly, and in a sense maintains its value estimate
implicitly through a mapping which signifies which path the agent
should take in order to maximize its average reward per stage.

\cite{MarTsi98} suggested an algorithm for non-discounted environments.
The gradient estimate is based on an estimate of the state values
which the actor estimates while interacting with the environment.
If the actor returns to a sequence of previously visited states, it
re-estimates the states value, not taking into account its previous
visits. This approach often results in large estimation variance.

\cite{BaxBar01} proposed an online algorithm for partially observable
MDPs. In this algorithm, the agent estimates the expected average
reward for the non-discounted problems through an estimate of the
value function of a related discounted problem. It was shown that
when the discount factor approaches $1$, the related discounted problem
approximates the average reward per stage. Similar to the algorithms
in (\cite{MarTsi98}), it suffers from relatively large estimation
variance. In (\cite{BaxBarGre04}), a method was proposed for coping
with the large variance by adding a baseline to the value function
estimation.

\subsection{Actor Critic Algorithms \label{sec:previous_AC}}

As stated in section \ref{sec:Introduction}, the convergence proofs
of which we are aware for AC algorithms are based on two time scale
stochastic approximation (\cite{Borkar1997}), where the actor is
assumed to operate on a time scale which is much slower than that
used by the critic.

\cite{KonBor99} suggested a set of AC algorithms. In two of their
algorithms (Algorithms 3 and 6), parametrized policy based actors
were used while the critic was based on a lookup table. Those algorithms
and their convergence proofs were specific to the Gibbs policy function
in the actor.

As far as we are aware, \cite{KonTsi03} provided the first convergence
proof for an AC algorithm based on function approximation. The information
passed from the critic to the actor is the critic's action-value function,
and the critic's basis functions, which are explicitly used by the
actor. They provided a convergence proof of their TD($\lambda$) algorithm
where $\lambda$ approaches $1$. A drawback of the algorithm is that
the actor and the critic must share the information regarding the
actor's parameters. This detailed information sharing is a clear handicap
in a biological context, which was one of the driving forces for the
present work.

Finally, \cite{BhaSutGhaLee08a,BhaSutGhaLee08b} recently proposed
an AC algorithm which closely resembles our proposed algorithm, and
which was developed independently of ours. In this work the actor
uses a parametrized policy function while the critic uses a function
approximation for the state evaluation. The critic passes to the actor
the TD(0) signal and based on it the actor estimates the average reward
gradient. A detailed comparison will be provided in section \ref{sec:comparison}.
As pointed out in \cite{BhaSutGhaLee08a,BhaSutGhaLee08b}, their work
is the first to provide a convergence proof for an AC algorithm incorporating
bootstrapping \cite{SutBar98}, where bootstrapping refers to a situation
where estimates are updated based on other estimates, rather than
on direct measurements (as in Monte Carlo approaches). This feature
applies to our work as well. We also note that \cite{BhaSutGhaLee08a,BhaSutGhaLee08b}
extend their approach to the so-called natural gradient estimator,
which has been shown to lead to improved convergence in supervised
learning as well as RL. The present study focuses on the standard
gradient estimate, leaving the extension to natural gradients to future
work.

\section{The Problem Setup \label{sec:ProblemSetup}}

In this section we describe the formal problem setup, and present
a sequence of assumptions and lemmas which will be used in order to
prove convergence of Algorithm 1 in section \ref{sec:MainAlgorithm}.
These assumptions and lemmas mainly concern the properties of the
controlled Markov chain, which represents the environment, and the
properties of the actor's parametrized policy function.

\subsection{The Dynamics of the Environment and of the Actor}

We consider an agent, composed of an actor and a critic, interacting
with an environment. We model the environment as a \emph{Markov Decision
Process} (MDP) \cite{Puterman1994} in discrete time with a finite
state set $\mathcal{X}$ and an action set $\mathcal{U}$, which may
be uncountable. \label{def:XS_US} We denote by $|\mathcal{X}|$ the
size of the set $\mathcal{X}$. Each selected action $u\in U$ determines
a stochastic matrix $P(u)=[P(y|x,u)]_{x,y\in\XS}$ \label{def:P(u)}
where $P(y|x,u)$ is the transition probability from a state $x\in\XS$
to a state $y\in\XS$ given the control $u$. For each state $x\in\XS$
the agent receives a corresponding reward $r(x)$, which may be deterministic
or random. In the present study we assume for simplicity that the
reward is deterministic, a benign assumption which can be easily generalized.
\begin{assumption}
\label{asum:r_bounded} The rewards, $\{r(x)\}_{x\in\XS}$, are uniformly
bounded by a finite constant $B_{r}$.
\end{assumption}
The actor maintains a \emph{parametrized policy function}. A parametrized
policy function is a conditional probability function, denoted by
$\mu(u|x,\theta)$, which maps an observation $x\in\XS$ into a control
$u\in\US$ given a parameter $\theta\in\mathbb{R}^{K}$. \label{def:theta}
The agent's goal is to adjust the parameter $\theta$ in order to
attain maximum average reward over time. For each $\theta$, we have
a Markov Chain (MC) induced by $P(y|x,u)$ and $\mu(u|x,\theta)$.
The state transitions of the MC are obtained by first generating an
action $u$ according to $\mu(u|x,\theta)$, and then generating the
next state according to $\{P(y|x,u)\}_{x,y\in\XS}$. Thus, the MC
has a transition matrix $P(\theta)=[P(y|x,\theta)]_{x,y\in\XS}$ which
is given by \begin{equation}
P(y|x,\theta)=\int_{\mathcal{U}}P(y|x,u)d\mu(u|x,\theta).\label{eq:trans_mat_policy}\end{equation}

We denote the space of these transition probabilities by $\PS=\{P(\theta)|\theta\in\R^{K}\}$,
and its closure by $\bar{\PS}$. \label{def:P(theta)} The following
assumption is needed in the sequel in order to prove the main results
(see \cite{Bremaud1999} for definitions).
\begin{assumption}
\label{asum:aperiodicy_reccurent} Each MC, $P(\theta)\in\bar{\PS}$,
is aperiodic, recurrent, and irreducible.
\end{assumption}
As a result of Assumption \ref{asum:aperiodicy_reccurent}, we have
the following lemma regarding the stationary distribution and a common
recurrent state.
\begin{lemma}
\label{lemma:ergodicy} Under Assumption \ref{asum:aperiodicy_reccurent}
we have: \end{lemma}
\begin{enumerate}
\item Each MC, $P(\theta)\in\bar{\PS}$, has a unique stationary distribution,
denoted by $\pi(\theta)$, satisfying $\pi(\theta)'P(\theta)=\pi(\theta)'$.
\item There exists a state, denoted by $x^{*}$, which is recurrent for
all $P(\theta)\in\bar{\PS}$. \end{enumerate}
\begin{proof}
For the first part see Corollary 4.1 in \citep{Gallager1995}. The
second part follows trivially from Assumption \ref{asum:aperiodicy_reccurent}.
\end{proof}
The next technical assumption states that the first and second derivatives
of the parametrized policy function are bounded, and is needed to
prove Lemma \ref{lemma:pi_eta_P_bounded} below.
\begin{assumption}
\label{asum:mu_bound} The conditional probability function $\mu(u|x,\theta)$
is twice differentiable. Moreover, there exist positive constants,
$B_{\mu_{1}}$ and $B_{\mu_{2}}$, such that for all $x\in\XS$, $u\in\US$,
$\theta\in\R^{K}$ and $k_{1}\ge1,\, k_{2}\le K$ we have \[
\left|\frac{\partial\mu(u|x,\theta)}{\partial\theta_{k}}\right|\le B_{\mu_{1}},\quad\left|\frac{\partial^{2}\mu(u|x,\theta)}{\partial\theta_{k_{1}}\partial\theta_{k_{2}}}\right|\le B_{\mu_{2}}.\]
\\
 \textbf{A notational comment concerning bounds} Throughout the
paper we denote upper bounds on different variables by the letter
$B$, with a subscript corresponding to the variable itself. An additional
numerical subscript, $1$ or $2$, denotes a bound on the first or
second derivative of the variable. For example, $B_{f}$, $B_{f_{1}}$,
and $B_{f_{2}}$ denote the bounds on the function $f$ and its first
and second derivatives respectively.
\end{assumption}

\subsection{Performance Measures}

Next, we define a performance measure for an agent in an environment.
The \textit{average reward per stage} of an agent which traverses
a MC starting from an initial state $x\in\mathcal{X}$ is defined
by \[
J(x,\theta)\triangleq\lim_{T\rightarrow\infty}\E\left[\left.\frac{1}{T}\sum_{n=0}^{T-1}r(x_{n})\right|x_{0}=x,\theta\right],\]
 where $\E[\cdot|\theta]$ denotes the expectation under the probability
measure $P(\theta)$, and $x_{n}$ is the state at time $n$. The
agent's goal is to find $\theta\in\R^{K}$ which maximizes $J(x,\theta)$.
The following lemma shows that under Assumption \ref{asum:aperiodicy_reccurent},
the average reward per stage does not depend on the initial state
(\cite{Ber2006}, vol. II, section 4.1).
\begin{lemma}
\label{lemma:starting_state} Under Assumption \ref{asum:aperiodicy_reccurent}
and based on Lemma \ref{lemma:ergodicy}, the average reward per stage,
$J(x,\theta)$, is independent of the starting state, is denoted by
$\eta(\theta)$, and satisfies $\eta(\theta)=\pi(\theta)'r$.
\end{lemma}
Based on Lemma \ref{lemma:starting_state}, the agent's goal is to
find a parameter vector $\theta$, which maximizes the average reward
per stage $\eta(\theta)$. In the sequel we show how this maximization
can be performed by optimizing $\eta(\theta)$, using $\nabla_{\theta}\eta(\theta)$.
A consequence of Assumption \ref{asum:mu_bound} and the definition
of $\eta(\theta)$ is the following lemma.
\begin{lemma}
\label{lemma:pi_eta_P_bounded} $ $ \end{lemma}
\begin{enumerate}
\item For each $x,y\in\XS$, $1\le i,j\le K$, and $\theta\in\R^{K}$, the
functions $\partial P(y|x,\theta)/\partial\theta_{i}$ and $\partial^{2}P(y|x,\theta)/\partial\theta_{i}\partial\theta_{j}$
are uniformly bounded by $B_{P_{1}}$ and $B_{P_{2}}$ respectively.

\begin{enumerate}
\item For each $x\in\XS$, $1\le i,j\le K$, and $\theta\in\R^{K}$, the
functions $\partial\pi(x|\theta)/\partial\theta_{i}$ and $\partial^{2}\pi(x|\theta)/\partial\theta_{i}\partial\theta_{j}$
are uniformly bounded by , $B_{\pi_{1}}$and $B_{\pi_{2}}$ respectively.
\item For all $1\le i,j\le K$, and $\theta\in\R^{K}$, the functions $\eta(\theta)$,
$\partial\eta(\theta)/\partial\theta_{i}$ and $\partial^{2}\pi(x|\theta)/\partial\theta_{i}\partial\theta_{j}$
are uniformly bounded by , $B_{\eta}$, $B_{\eta_{1}}$ and $B_{\eta_{2}}$
respectively.
\item For all $x\in\XS$ and $\theta\in\R^{K}$, there exists a constant
$b_{\pi}>0$ such that $\pi(x|\theta)\ge b_{\pi}$.
\end{enumerate}
\end{enumerate}
The proof is technical and is given in Appendix \ref{app:pi_eta_P_bounded}.
For later use, we define the random variable $T$, which denotes the
first return time to the recurrent state $x^{*}$. Formally, \begin{equation}
T\triangleq\min\{k>0|x_{0}=x^{*},\, x_{k}=x^{*}\}.\end{equation}
 It is easy to show that under Assumption \ref{asum:aperiodicy_reccurent},
the average reward per stage can be expressed by \begin{equation}
\eta(\theta)=\lim_{T\rightarrow\infty}\E\left[\left.\frac{1}{T}\sum_{n=0}^{T-1}r(x_{n})\right|x_{0}=x^{*},\theta\right].\end{equation}
 Next, we define the \textit{differential value function} of state
$x\in\XS$ which represents the average differential reward the agent
receives upon starting from a state $x$ and reaching the recurrent
state $x^{*}$ for the first time. Mathematically, \begin{equation}
h(x,\theta)\triangleq\E\left[\left.\sum_{n=0}^{T-1}(r(x_{n})-\eta(\theta))\right|x_{0}=x,\theta\right].\label{eq:h_def}\end{equation}
 Abusing notation slightly, we denote $h(\theta)\triangleq(h(x_{1},\theta),\ldots,h(x_{|\XS|},\theta))\in\R^{|\XS|}$.
For each $\theta\in\R^{K}$ and $x\in\XS$, $h(x,\theta)$, $r(x)$,
and $\eta(\theta)$ satisfy Poisson's equation (see Theorem 7.4.1
in (\cite{Ber2006})), i.e., \begin{equation}
h(x,\theta)=r(x)-\eta(\theta)+\sum_{y\in\XS}P(y|x,\theta)h(y,\theta).\label{eq:Poisson_eq}\end{equation}
Based on the differential value we define the \textit{temporal difference}
(TD) between the states $x\in\XS$ and $y\in\XS$ (see \cite{BerTsi96},
\cite{SutBar98}), \begin{equation}
d(x,y,\theta)\triangleq r(x)-\eta(\theta)+h(y,\theta)-h(x,\theta).\label{eq:TD_def}\end{equation}
According to common wisdom, the TD is interpreted as a prediction
error. The next lemma states the boundedness of $h(x,\theta)$ and
its derivatives. The proof is given in Appendix \ref{app:h_bounded}.
\begin{lemma}
\label{lemma:h_bounded}$ $\end{lemma}
\begin{enumerate}
\item The differential value function, $h(x,\theta)$, is bounded and has
bounded first and second derivative. Mathematically, for all $x\in\XS$,
$1\le i,j\le K$, and for all $\theta\in\R^{K}$ we have \[
\left|h(x,\theta)\right|\le B_{h},\quad\left|\frac{\partial h(x,\theta)}{\partial\theta_{i}}\right|\le B_{h_{1}},\quad\left|\frac{\partial^{2}h(x,\theta)}{\partial\theta_{i}\partial\theta_{j}}\right|\le B_{h_{2}}.\]

\begin{enumerate}
\item There exists a constant $B_{D}$ such that or all $\theta\in\R^{K}$
we have $\left|d(x,y,\theta)\right|\le B_{D}$, where $B_{D}=2\left(B_{r}+B_{h}\right)$.
\end{enumerate}
\end{enumerate}

\subsection{The Critic's Dynamics}

\begin{figure}[hbt]
\begin{centering}
\hspace*{-.5cm} \vspace{-0.5cm}
 \includegraphics[scale=0.5]{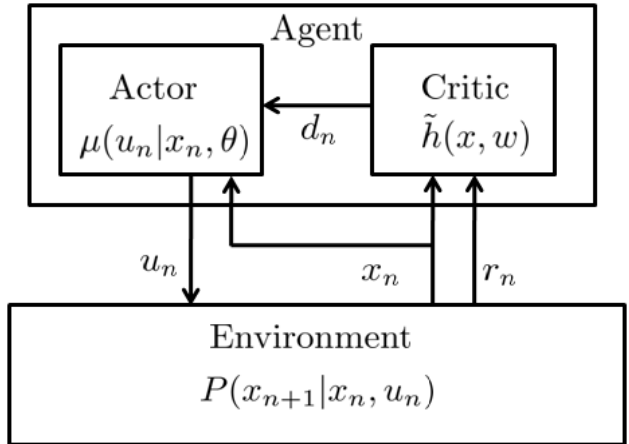}
\par\end{centering}

\caption{A schematic illustration of the dynamics between the actor, the critic,
and the environment. The actor chooses an action, $u_{n}$, according
to the parametrized policy $\mu(u|x,\theta)$. As a result, the environment
proceeds to the next state according to the transition probability
$P(x_{n+1}|x_{n},u_{n})$ and provides a reward. Using the TD signal,
the critic improves its estimation for the environment state values
while the actor improves its policy. \label{fig:AC_Diagram.ps} }

\end{figure}

The critic maintains an estimate of the environmental state values.
It does so by maintaining a parametrized function which approximates
$h(x,\theta)$, and is denoted by $\he(x,w)$. The function $\he(x,w)$
is a function of the state $x\in\XS$ and a parameter $w\in\R^{L}$.
We note that $h(x,\theta)$ is a function of $\theta$, and is induced
by the actor policy $\mu(u|x,\theta)$, while $\he(x,w)$ is a function
of $w$. Thus, the critic's objective is to find the parameter $w$
which yields the best approximation of $h(\theta)=(h(x_{1},\theta),\ldots,h(x_{|\XS|},\theta))$,
in a sense to be defined later. We denote this optimal vector by $w^{*}(\theta)$.
An illustration of the interplay between the actor, critic, and the
environment is given in Figure \ref{fig:AC_Diagram.ps}.\\

\section{A Single Time Scale Actor Critic Algorithm with Linear Function Approximation
\label{sec:MainAlgorithm}}

In this section, we present a version of an AC algorithm, along with
its convergence proof. The core of the algorithm is based on \eqref{eq:grad_eta_based TD}
below, where the actor's estimate of $\nabla_{\theta}\eta(\theta)$
is based on the critic's estimate of the TD signal $d(x,y,\theta)$.
The algorithm is composed of three iterates, one for the actor and
two for the critic. The actor maintains the iterate of the parameter
vector $\theta$ corresponding to the policy $\mu(u|x,\theta)$, where
its objective is to find the optimal value of $\theta$, denoted by
$\theta^{*}$, which maximizes $\eta(\theta)$. The critic maintains
the other two iterates. One iterate is used for estimating the average
reward per stage, $\eta(\theta)$, where its estimate is denoted by
$\ee$. The critic's second iterate maintains a parameter vector,
denoted by $w\in\R^{L}$, which is used for the differential value
estimate using a function approximator, denoted by $\he(w)$. For
each $\theta\in\R^{K}$, there exists a $w^{*}(\theta)$ which, under
the policy induced by $\theta$, is the optimal $w$ for estimating
$\ee(w)$. The critic's objective is to find the optimal $\ee$ and
$w$.

\subsection{Using the TD Signal to Estimate the Gradient of the Average Reward
\label{sec:Using_TD_Signal}}

We begin with a theorem which serves as the foundation for the policy
gradient algorithm described in Section \ref{sec:MainAlgorithm}.
The theorem relates the gradient of the average reward per stage,
$\eta(\theta)$, to the TD signal. It was proved in (\cite{BhaSutGhaLee08a}),
and is similar in its structure to other theorems which connect $\eta(\theta)$
to the $Q$-value (\cite{KonTsi03}), and to the differential value
function (\cite{Cao07,MarTsi98}).

We start with a definition of the \textit{likelihood ratio derivative}
\[
\psi(x,u,\theta)\triangleq\frac{\nt\mu(u|x,\theta)}{\mu(u|x,\theta)},\]
where the gradient $\nt$ is w.r.t.\! $\theta$, and $\psi(x,u,\theta)\in\R^{K}$.
The following assumption states that $\psi(x,u,\theta)$ is bounded,
and will be used to prove the convergence of algorithm \ref{algo:TD1ACFunction}.
\begin{assumption}
\label{asum:psi_bounded} For all $x\in\XS$, $u\in\US$, and $\theta\in\R^{K}$,
there exists a positive constant, $B_{\psi}$, such that \[
\left\Vert \psi(x,u,\theta)\right\Vert _{2}\le B_{\psi}<\infty,\]
 where $\|\cdot\|_{2}$ is the Euclidean $L_{2}$ norm.
\end{assumption}
Based on this, we present the following theorem which relates the
gradient of $\eta(\theta)$ to the TD signal. For completeness, we
supply a (straightforward) proof in Appendix \ref{app:grad_eta_based_TD}.
\begin{theorem}
\label{thrm:grad_eta_based TD} For any arbitrary function $f(x)$,
the gradient w.r.t. $\!\theta$ of the average reward per stage can
be expressed by \begin{equation}
\nt\eta(\theta)=\sum_{x,y\in\mathcal{X}}P(x,u,y,\theta)\psi(x,u,\theta)d(x,y,\theta),\label{eq:grad_eta_based TD}\end{equation}
 where $P(x,u,y,\theta)$ is the probability $\Pr(x_{n}=x,u_{n}=u,x_{n+1}=y)$
subject to the policy parameter $\theta$.
\end{theorem}

\subsection{The updates performed by the critic and the actor }

We note that the following derivation regarding the critic is similar
in some respects to the derivation in section 6.3.3 of \cite{BerTsi96}
and of \cite{TsitsiklisVanRoy1997}. We define the following quadratic
target function used to evaluate the critic's performance in assessing
the differential value $h(\theta)$, \begin{equation}
I(w,\theta)\triangleq\frac{1}{2}\sum_{x\in\XS}\pi(x|\theta)\left(\he(x,w)-h(x,\theta)\right)^{2}~.\label{eq:theoretic_cost_function}\end{equation}
 The probabilities $\{\pi(x|\theta)\}_{x\in\XS}$ are used in order
to provide the proportional weight to the state estimates, according
to the relative number of visits of the agent to the different states.

Limiting ourselves to the class of linear function approximations
in the critic, we consider the following function for the differential
value function \begin{equation}
\he(x,w)=\phi(x)'w,\label{eq:h_scalar_linear_approx_def}\end{equation}
 where $\phi(x)\in\R^{L}$. We define $\Phi\in\R^{|\XS|\times L}$
to be the matrix \[
\Phi\triangleq\left(\begin{array}{cccc}
\phi_{1}(x_{1}) & \phi_{2}(x_{1}) & \ldots & \phi_{L}(x_{1})\\
\phi_{1}(x_{2}) & \phi_{2}(x_{2}) & \ldots & \phi_{L}(x_{2})\\
\vdots & \vdots &  & \vdots\\
\phi_{1}(x_{|\XS|}) & \phi_{2}(x_{|\XS|}) & \ldots & \phi_{L}(x_{|\XS|})\end{array}\right)~,\]
 where $\phi(\cdot)$ is a column vector. Therefore, we can express
\eqref{eq:h_scalar_linear_approx_def} in vector form as \begin{equation}
\he(w)=\Phi w,\label{eq:H_eq_phi_w}\end{equation}
 where, abusing notation slightly, we set $\tilde{h}(w)=\left(\tilde{h}(x_{1},w),\ldots,\tilde{h}(x_{\left|\XS\right|},w)\right)'$.
$ $

We wish to express \eqref{eq:theoretic_cost_function}, and the approximation
process, in an appropriate Hilbert space. Define the matrix $\Pi(\theta)$
to be a diagonal matrix $\Pi(\theta)\triangleq\textrm{diag}(\pi(\theta))$.
Thus, \eqref{eq:theoretic_cost_function} can be expressed as \begin{equation}
I(w,\theta)=\frac{1}{2}\left\Vert \Pi(\theta)^{\frac{1}{2}}\left(h(\theta)-\Phi w\right)\right\Vert _{2}^{2}\triangleq\frac{1}{2}\left\Vert h(\theta)-\Phi w\right\Vert _{\Pi(\theta)}^{2}.\label{eq:cost_function_vector_form}\end{equation}
 In the sequel, we will need the following technical assumption.
\begin{assumption}
$\quad$\label{asum:Phi_linear_indepen_bound}\end{assumption}
\begin{enumerate}
\item The columns of the matrix $\Phi$ are independent, i.e., they form
a basis of dimension $L$.

\begin{enumerate}
\item The norms of the column vectors of the matrix $\Phi$ are bounded
above by $1$, i.e., $\|\phi_{k}\|_{2}\le1$ for $1\le k\le L$.
\end{enumerate}
\end{enumerate}
The parameter $w^{*}(\theta)$, which optimizes \eqref{eq:cost_function_vector_form},
can be directly computed, but involves inverting a matrix. Thus, in
order to find the right estimate for $\he(w)$, the following \emph{gradient
descent} (\cite{BerTsi96}) algorithm is suggested, \begin{equation}
w_{n+1}=w_{n}-\gamma_{n}\nw I(w_{n},\theta),\label{eq:w_m_def}\end{equation}
 where $\{\gamma_{n}\}_{n=1}^{\infty}$ is a positive series satisfying
the following assumption, which will be used in proving the convergence
of Algorithm 1.
\begin{assumption}
\label{ass:gamma_summability}The positive series $\{\gamma_{n}\}_{n=1}^{\infty}$
satisfies \begin{equation}
\sum_{n=1}^{\infty}\gamma_{n}=\infty,\quad\sum_{n=1}^{\infty}\gamma_{n}^{2}<\infty.\label{eq:gamma_coeff}\end{equation}

\end{assumption}
Writing the term $\nw I(w_{n})$ explicitly yields \begin{equation}
\nw I(w_{n})=\Phi'\Pi(\theta)\Phi w_{n}-\Phi'\Pi(\theta)h(\theta).\label{eq:w_nabla_explicit}\end{equation}
 For each $\theta\in\R^{K}$, the value $w^{*}(\theta)$ is given
by setting $\nw I(w,\theta)=0$, i.e.,

\begin{equation}
w^{*}(\theta)=\left(\Phi'\Pi(\theta)\Phi\right)^{-1}\Phi'\Pi(\theta)h(\theta).\label{eq:w_star_explicit}\end{equation}
Note that \cite{BerTsi96} prove that the matrix $\left(\Phi'\Pi(\theta)\Phi\right)^{-1}\Phi'\Pi(\theta)$
is a projection operator into the space spanned by $\Phi w$, with
respect to the norm $\left\Vert \cdot\right\Vert _{\Pi\left(\theta\right)}$
. Thus, the explicit gradient descent procedure \eqref{eq:w_m_def}
is \begin{equation}
\begin{split}w_{n+1} & =w_{n}-\gamma_{n}\Phi'\Pi\left(\theta\right)\left(\Phi w_{n}-h(\theta)\right).\end{split}
\label{eq:w_m_def_explicit}\end{equation}
Using the basis $\Phi$, in order to approximates $h\left(\theta\right)$,
yields an approximation error defined by\[
\epsilon_{\textrm{app}}\left(\theta\right)\triangleq\inf_{w\in\mathbb{R}^{L}}\left\Vert h\left(\theta\right)-\Phi w\right\Vert _{\pi\left(\theta\right)}=\left\Vert h\left(\theta\right)-\Phi w^{*}\left(\theta\right)\right\Vert _{\pi\left(\theta\right)}.\]
We can bound this error by\begin{equation}
\epsilon_{\textrm{app}}\triangleq\sup_{\theta\in\mathbb{R}^{K}}\epsilon_{\textrm{app}}\left(\theta\right).\label{eq:FA_err_bound}\end{equation}

The agent cannot access $h(x,\theta)$ directly. Instead, it can interact
with the environment in order to estimate $h(x,\theta)$. We denote
by $\hat{h}_{n}(x)$ the estimate of $h(x,\theta)$ at time step $n$,
thus \eqref{eq:w_m_def_explicit} becomes \begin{equation}
w_{n+1}=w_{n}+\gamma_{n}\Phi'\Pi(\theta)\left(\hat{h}_{n}-\Phi w_{n}\right).\label{eq:C_suggestion}\end{equation}
This procedure is termed \emph{stochastic gradient descent} (\cite{BerTsi96}).

There exist several estimators for $\hat{h}_{n}$. One sound method,
which performs well in practical problems (see \cite{Tesauro1995}),
is the TD($\lambda)$ method (see section 5.3.2 and 6.3.3 in \cite{BerTsi96},
or Chapter 6 in \cite{SutBar98}), where the parameter $\lambda$
satisfies $0\le\lambda\le1$. This method devises an estimator which
is based on previous estimates of $h\left(w\right)$, i.e., $w_{n}$,
and is based also on the environmental reward $r\left(x_{n}\right)$.
This idea is a type of \emph{a bootstrapping} algorithm, i.e., using
existing estimates and new information in order to build more accurate
estimates (see \cite{SutBar98}, Section 6.1).

The TD$\left(\lambda\right)$ estimator for $\hat{h}_{n+1}$ is \begin{eqnarray}
\hat{h}_{n+1}\left(x_{n}\right) & = & \left(1-\lambda\right)\sum_{k=0}^{\infty}\lambda^{k}\hat{h}_{n+1}^{\left(k\right)}\left(x_{n}\right),\label{eq:TD_naive_predictor}\end{eqnarray}
where the\emph{ $k$-steps predictor} is defined by\[
\hat{h}_{n+1}^{\left(k\right)}\left(x_{n}\right)=\left(\sum_{m=0}^{k}r\left(x_{n+m}\right)+\hat{h}_{n}\left(x_{n+k+1}\right)\right).\]
The idea of bootstrapping is apparent in \eqref{eq:TD_naive_predictor}:
the predictor for the differential value of the state $x_{n}$ at
the $\left(n+1\right)$-Th time step, is based partially on the previous
estimates through $\hat{h}_{n}\left(x_{n+k+1}\right)$, and partially
on new information, i.e., the reward $r\left(x_{n+m}\right)$. In
addition, the parameter $\lambda$ gives an exponential weighting
for the different $k$-step predictors. Thus, choosing the right $\lambda$
can yield better estimators.

For the discounted setting, it was proved by \cite{BerTsi96} (p.
295) that an algorithm which implements the TD$\left(\lambda\right)$
estimator \eqref{eq:TD_naive_predictor} online and converges to the
right value is the following one\begin{eqnarray}
w_{n+1} & = & w_{n}+\gamma_{n}d_{n}e_{n},\nonumber \\
e_{n} & = & \alpha\lambda e_{n-1}+\phi\left(x_{n}\right),\label{eq:discounted_critic}\end{eqnarray}
where $d_{n}$ is the temporal difference between the $n$-th and
the $\left(n+1\right)$-th cycle, and $e_{n}$ is the so-called \emph{eligibility
trace} (see Sections 5.3.3 and 6.3.3 in\cite{BerTsi96} or Chapter
7 in \cite{SutBar98}), and the parameter $\alpha$ is the discount
factor. The eligibility trace is an auxiliary variable, which is used
in order to implement the idea of \eqref{eq:TD_naive_predictor} as
an online algorithm. As the name implies, the eligibility variable
measures how eligible is the TD variable, $d_{n}$, in \eqref{eq:discounted_critic}.

In our setting, the non-discounted case, the analogous equations for
the critic, are \begin{eqnarray}
w_{n+1} & = & w_{n}+\gamma_{n}\tilde{d}\left(x_{n},x_{n+1},w_{n}\right)e_{n}\nonumber \\
\tilde{d}\left(x_{n},x_{n+1},w_{n}\right) & = & r(x_{n})-\ee_{m}+\he(x_{n+1},w_{m})-\he(x_{n},w_{m})\label{eq:critic_iterate_suggested}\\
e_{n} & = & \lambda e_{n-1}+\phi\left(x_{n}\right).\nonumber \end{eqnarray}

The actor's iterate is motivated by Theorem \ref{thrm:grad_eta_based TD}.
Similarly to the critic, the actor executes a stochastic gradient
ascent step in order to fior with a parametrized policy $\mu(u|x,\theta)$
satisfying Assumptions \ref{asum:mu_bound} and \ref{asum:psi_bounded}.
\begin{itemize}
\item A critic with
\end{itemize}
nd a local maximum of the average reward per stage $\eta(\theta)$.
Therefore, \begin{equation}
\theta_{n+1}=\theta_{n}+\gamma_{n}\psi(x_{n},u_{n},\theta_{n})\de_{n}(x_{n},x_{n+1},w_{n}).\label{eq:A_suggestion}\end{equation}
A summary of the algorithm is presented in Algorithm \ref{algo:TD1ACFunction}.

\begin{algorithm}
Given:
\begin{itemize}
\item An MDP with a finite set $\XS$ of states satisfying Assumption \ref{asum:aperiodicy_reccurent}.
\item An actor with a parametrized policy $\mu(u|x,\theta)$ satisfying
Assumptions \ref{asum:mu_bound} and \ref{asum:psi_bounded}.
\item A critic with a linear basis for $\he(w)$, i.e., $\{\phi\}_{i=1}^{L}$,
satisfying Assumption \ref{asum:Phi_linear_indepen_bound}.
\item A set $H$, a constant $B_{w}$, and an operator $\Psi_{w}$ according
to Definition \ref{asum:theta_w_constrained}.
\item Step parameters $\Gamma_{\eta}$ and $\Gamma_{h}$.
\item Choose a TD parameter $0\le\lambda<1$.
\end{itemize}
For step $n=0:$
\begin{itemize}
\item Initiate the critic and the actor variables: $\ee_{0}=0$ ,$w_{0}=0$,
$e_{0}=0$, $\theta_{0}=0$.
\end{itemize}
For each step $n=1,2,\ldots$

\textbf{$\quad$Critic:} Calculate the estimated TD and eligibility
trace \begin{eqnarray}
\tilde{\eta}{}_{n+1} & = & \tilde{\eta}_{n}+\gamma_{n}\Gamma_{\eta}\left(r(x_{n})-\tilde{\eta}_{n}\right)\label{eq:eta_iterate}\\
\he(x,w_{n}) & = & w_{n}'\phi(x),\nonumber \\
\tilde{d}\left(x_{n},x_{n+1},w_{n}\right) & = & r(x_{n})-\ee_{n}+\he(x_{n+1},w_{n})-\he(x_{n},w_{n}),\nonumber \\
e_{n} & = & \lambda e_{n-1}+\phi\left(x_{n}\right).\nonumber \end{eqnarray}
\textbf{$\quad$} Set,\begin{eqnarray}
w_{n+1} & = & w_{n}+\gamma_{n}\Gamma_{w}\tilde{d}Cramer's\left(x_{n},x_{n+1},w_{n}\right)e_{n}\label{eq:w_iterate}\end{eqnarray}
\textbf{$\quad$} \textbf{Actor:} \begin{equation}
\theta_{n+1}=\theta_{n}+\gamma_{n}\psi(x_{n},u_{n},\theta_{n})\de_{n}(x_{n},x_{n+1},w_{n})\label{eq:theta_w_iterate}\end{equation}
\textbf{$\quad$}Project each component of $w_{m+1}$ onto $H$ (see
Definition \ref{asum:theta_w_constrained})

\caption{TD AC Algorithm\label{algo:TD1ACFunction}}

\end{algorithm}

\subsection{Convergence Proof for the AC Algorithm}

In the remainder of this section, we state the main theorems related
to the convergence of Algorithm \ref{algo:TD1ACFunction}. We present
a sketch of the proof in this section, where the technical details
are relegated to Appendices \ref{app:main1_algoritm_to_ODE} and \ref{app:main2_ODE_convergence}.
The proof is divided into two stages. In the first stage we relate
the stochastic approximation to a set of ordinary differential equations
(ODE). In the second stage, we find conditions under which the ODE
system converges to a neighborhood of the optimal $\eta(\theta)$.

The ODE approach is a widely used method in the theory of stochastic
approximation for investigating the asymptotic behavior of stochastic
iterates, such as \eqref{eq:eta_iterate}-\eqref{eq:theta_w_iterate}.
The key idea of the technique is that the iterate can be decomposed
into a mean function and a noise term, such as a martingale difference
noise. As the iterates advance, the effect of the noise weakens due
to repeated averaging. Moreover, since the step size of the iterate
decreases (e.g., $\gamma_{n}$ in \eqref{eq:eta_iterate}-\eqref{eq:theta_w_iterate}),
one can show that asymptotically an interpolation of the iterates
converges to a continuous solution of the ODE. Thus, the first part
of the convergence proof is to find the ODE system which describes
the asymptotic behavior of Algorithm 1. This ODE will be presented
in Theorem \ref{thrm:main1_algoritm_to_ODE}. In the second part we
use ideas from the theory of Lyapunov functions in order to characterize
the relation between the constants, $|\XS|$, $\Gamma_{\eta}$, $\Gamma_{w}$,
etc., which ensure convergence to some neighborhood of the maximum
point satisfying $\|\nt\eta(\theta)\|_{2}=0$. Theorem \ref{thrm:main2_ODE_convergence}
states conditions on this convergence.

\subsubsection{Relate the Algorithm to an ODE}

In order to prove the convergence of this algorithm to the related
ODE, we need to introduce the following assumption, which adds constraints
to the iteration for $w$, and will be used in the sequel to prove
Theorem \ref{thrm:main1_algoritm_to_ODE}. This assumption may seem
restrictive at first but in practice it is not. The reason is that
we usually assume the bounds of the constraints to be large enough
so the iterates practically do not reach those bounds. For example,
under Assumption \ref{asum:aperiodicy_reccurent} and additional mild
assumptions, it is easy to show that $h(\theta)$ is uniformly bounded
for all $\theta\in\R^{K}$. As a result, there exist a constant bounding
$w^{*}(\theta)$ for all $\theta\in\R^{K}$. Choosing constraints
larger than this constant will not influence the algorithm performance.
\begin{definition}
\label{asum:theta_w_constrained} Let us denote by $\{w_{i}\}_{i=1}^{L}$
the components of $w$, and choose a positive constant $B_{w}$. We
define the set $H\subset\R^{K}\times\R^{L}$ to be \[
H\triangleq\left\{ (\theta,w)\left|-\infty<\theta_{i}<\infty,\quad1\le i\le K,\quad-B_{w}\le w_{j}\le B_{w},\quad1\le j\le L\right.\right\} ,\]
 and let $\Psi_{w}$ be an operator which projects $w$ onto $H$,
i.e., for each $Cramer's1\le j\le L$, $\Psi_{w}w_{j}=\max(\min(w_{j},B_{w}),-B_{w})$.
\end{definition}
The following theorem identifies the ODE system which corresponds
to Algorithm \ref{algo:TD1ACFunction}. The detailed proof is given
in Appendix \ref{app:main1_algoritm_to_ODE}.
\begin{theorem}
\label{thrm:main1_algoritm_to_ODE} Define the following functions:

\begin{eqnarray}
G(\theta) & = & \Phi'\Pi\left(\theta\right)\sum_{m=0}^{\infty}\lambda^{m}P\left(\theta\right)^{m},\nonumber \\
D^{(x,u,y)}(\theta) & = & \pi\left(x\right)P\left(u|x,\theta\right)P\left(y|x,u\right)\psi\left(x,u,\theta\right),\quad x,y\in\XS,\quad u\in\US.\label{eq:thrm_main_vars_w}\\
A\left(\theta\right) & = & \Phi'\Pi\left(\theta\right)\left(M\left(\theta\right)-I\right)\Phi,\nonumber \\
M\left(\theta\right) & = & \left(1-\lambda\right)\sum_{m=0}^{\infty}\lambda^{m}P\left(\theta\right)^{m+1},\nonumber \\
b\left(\theta\right) & = & \Phi'\Pi\left(\theta\right)\sum_{m=0}^{\infty}\lambda^{m}P\left(\theta\right)^{m}\left(r-\eta\left(\theta\right)\right).\nonumber \end{eqnarray}
Then, \end{theorem}
\begin{enumerate}
\item Algorithm 1 converges to the invariant set of the following set of
ODEs \begin{equation}
\left\{ \begin{split}\dot{\theta}= & \nt\eta(\theta)+\sum_{x,y\in\XS\times\XS}D^{(x,u,y)}(\theta)\left(d(x,y,\theta)-\tilde{d}(x,y,w)\right),\\
\dot{w}= & \Psi_{w}\left[\Gamma_{w}\left(A\left(\theta\right)w+b\left(\theta\right)+G(\theta)(\eta(\theta)-\ee)\right)\right],\\
\dot{\ee}= & \Gamma_{\eta}\left(\eta(\theta)-\ee\right),\end{split}
\right.\label{eq:ODE_w}\end{equation}
 with probability 1.

\begin{enumerate}
\item The functions in \eqref{eq:thrm_main_vars_w} are continuous with
respect to $\theta$.
\end{enumerate}
\end{enumerate}

\subsubsection{Investigating the ODE Asymptotic Behavior}

Next, we quantify the asymptotic behavior of the system of ODEs in
terms of the various algorithmic parameters. The proof of the theorem
appears in Appendix \ref{app:main2_ODE_convergence}.
\begin{theorem}
\label{thrm:main2_ODE_convergence} Consider the constants $\Gamma_{\eta}$
and $\Gamma_{w}$ as defined in Algorithm 1, and the function approximation
bound $\epsilon_{\textrm{app}}$ as defined in \eqref{eq:FA_err_bound}.
Setting \[
B_{\nabla\eta}\triangleq\frac{B_{\Delta td1}}{\Gamma_{w}}+\frac{B_{\Delta td2}}{\Gamma_{\eta}}+B_{\Delta td3}\epsilon_{\textrm{\textrm{app}}},\]
 where $B_{\Delta td1}$, $B_{\Delta td2}$, $B_{\Delta td3}$ are
a finite constants depending on the MDP and agent parameters. Then,
the ODE system \eqref{eq:ODE_w} satisfies \begin{equation}
\liminf_{t\rightarrow\infty}\|\nt\eta(\theta_{t})\|\le\Bgradeta.\end{equation}

\end{theorem}
Theorem \ref{thrm:main2_ODE_convergence} has a simple interpretation.
Consider the trajectory $\eta(\theta_{t})$ for large times, corresponding
to the asymptotic behavior of $\eta_{n}$. The result implies that
the trajectory visits a neighborhood of a local maximum infinitely
often. Although it may leave the local vicinity of the maximum, it
is guaranteed to return to it infinitely often. This occurs, since
once it leaves the vicinity, the gradient of $\eta$ points in a direction
which has a positive projection on the gradient direction, thereby
pushing the trajectory back to the vicinity of the maximum. It should
be noted that in simulation (reported below) the trajectory usually
remains within the vicinity of the local maximum, rarely leaving it.
We also observe that by choosing appropriate values for $\Gamma_{\eta}$
and $\Gamma_{w}$ we can control the size of the ball to which the
algorithm converges.

The key idea required to prove the Theorem is the following argument.
If the trajectory does not satisfy $\|\nabla\eta(\theta)\|_{2}\le\Bgradeta$,
we have $\dot{\eta}(\theta)>\epsilon$ for some positive $\epsilon$.
As a result, we have a monotone function which increases to infinity,
thereby contradicting the boundedness of $\eta(\theta)$. Thus, $\eta(\theta)$
must visit the set which satisfies $\|\nabla\eta(\theta)\|_{2}\le\Bgradeta$
infinitely often.

\section{A Comparison to other convergence results\label{sec:comparison}}

In this section, we point out the main differences between Algorithm
\ref{algo:TD1ACFunction}, the first algorithm proposed by \cite{BhaSutGhaLee08b}
and the algorithms proposed by \cite{KonTsi03}. The main dimensions
along which we compare the algorithms are the time scale, the type
of the TD signal, and whether the algorithm is on line or off line.

\subsection*{The Time Scale and Type of Convergence}

As was mentioned previously, the algorithms of \cite{BhaSutGhaLee08b}
and \cite{KonTsi03} need to operate in two time scales. More precisely,
this refers to the following situation. Denote the time step of the
critic's iteration by $\gamma_{n}^{c}$ and the time step of the actor's
iteration by $\gamma_{n}^{a}$, we have $\gamma_{n}^{c}=o(\gamma_{n}^{a})$,
i.e., \[
\lim_{n\rightarrow\infty}\frac{\gamma_{n}^{c}}{\gamma_{n}^{a}}=0.\]
 The use of two time scales stems from the need of the critic to give
an accurate estimate of the state values (as in the work of \cite{BhaSutGhaLee08b})
or the state-action values (as in the work of \cite{KonTsi03}) before
the actor uses them.

In the algorithm proposed here, a single time scale is used for the
three iterates of Algorithm \ref{algo:TD1ACFunction}. We have $\gamma_{n}^{a}=\gamma_{n}$
for the actor iterate, $\gamma_{n}^{c,\eta}=\Gamma_{\eta}\gamma_{n}$
for the critic's $\eta_{n}$ iterate, and $\gamma_{n}^{c,w}=\Gamma_{w}\gamma_{n}$
for the critic's $w$ iterate. Thus, \[
\begin{split}\lim_{n\rightarrow\infty}\frac{\gamma_{n}^{c,\eta}}{\gamma_{n}^{a}} & =\Gamma_{\eta},\\
\lim_{n\rightarrow\infty}\frac{\gamma_{n}^{c,w}}{\gamma_{n}^{a}} & =\Gamma_{w}.\end{split}
\]

Due to the single time scale, Algorithm \ref{algo:TD1ACFunction}
has\textcolor{magenta}{ }the potential to converge faster than algorithms
based on two time scales, since both the actor and the critic may
operate on the fast time scale. The drawback of Algorithm \ref{algo:TD1ACFunction}
is the fact that convergence to the optimal value cannot be guaranteed,
as was proved by \cite{BhaSutGhaLee08b} and by \cite{KonTsi03}.
Instead, convergence to a neighborhood in $\R^{K}$ around the optimal
value is guaranteed. In order to make the neighborhood smaller, we
need to choose $\Gamma_{\eta}$ and $\Gamma_{w}$ appropriately, as
is stated in Theorem \ref{thrm:main2_ODE_convergence}.

\subsection*{The TD Signal, the Information Passed Between the Actor and the Critic,
and the Critic's Basis \label{sec:compare_algorithms}}

The algorithm presented in \cite{BhaSutGhaLee08b} is essentially
a TD(0) algorithm, while the algorithm in \cite{KonTsi03} is TD(1),
Our algorithm is a TD$\left(\lambda\right)$ for $0\le\lambda<1$.
A major difference between the approaches in \cite{BhaSutGhaLee08b}
and the present work, as compared to \cite{KonTsi03}, is the information
passed from the critic to the actor. In the former cases, the information
passed is the TD signal, while in the latter case the Q-value is passed.
Additionally, in \cite{BhaSutGhaLee08b} and in Algorithm \ref{algo:TD1ACFunction}
the critic's basis functions do not change through the simulation,
while in \cite{KonTsi03} the critic's basis functions are changed
in each iteration according to the actor's parameter $\theta$. Finally,
we comment that \cite{BhaSutGhaLee08b} introduced an additional algorithm,
based on the so-called natural gradient, which led to improved convergence
speed. In this work we limit ourselves to algorithms based on the
regular gradient, and defer the incorporation of the natural gradient
to future work. As stated in Section \ref{sec:Introduction}, our
motivation in this work was the derivation of a single time scale
online AC algorithm with guaranteed convergence, which may be applicable
in a biological context. The more complex natural gradient approach
seems more restrictive in this setting.

\section{Simulations}

\label{sec:simulations}

We report empirical results applying Algorithm \ref{algo:TD1ACFunction}
to a set of abstract randomly constructed MDPs which are termed Average
Reward Non-stationary Environment Test-bench or in short \textsc{garnet}
(\cite{Archibald1995}). \textsc{garnet} problems comprise a class
of randomly constructed finite MDPs serving as a test-bench for control
and RL algorithms optimizing the average reward per stage. A \textsc{garnet}
problem is characterized in our case by four parameters and is denoted
by \textsc{garnet}$(X,U,B,\sigma)$. The parameter $X$ is the number
of states in the MDP, $U$ is the number of actions, $B$ is the branching
factor of the MDP, i.e., the number of non-zero entries in each line
of the MDP's transition matrices, and $\sigma$ is the variance of
each transition reward.

We describe how a \textsc{garnet} problem is generated. When constructing
such a problem, we generate for each state a reward, distributed normally
with zero mean and unit variance. For each state-action the reward
is distributed normally with the state's reward as mean and variance
$\sigma^{2}$. The transition matrix for each action is composed of
$B$ non-zero terms in each line which sum to one.

We note that a comparison was carried out by \cite{BhaSutGhaLee08b}
between their algorithm and the algorithm of \cite{KonTsi03}. We
therefore compare our results directly to the more closely related
former approach (see also Section \ref{sec:compare_algorithms}).

We consider the same \textsc{garnet} problems as those simulated by
\cite{BhaSutGhaLee08b}. For completeness, we provide here the details
of the simulation. For the critic's feature vector, we use a linear
function approximation $\he(x,w)=\phi(x)'w$, where $\phi(x)\in\{0,1\}^{L}$,
and define $l$ to be the number nonzero values in $\phi(x)$. The
nonzero values are chosen uniformly at random, where any two states
have different feature vectors. The actor's feature vectors are of
size $L\times|\US|$, and are constructed as \begin{align*}
\xi(x,u) & \triangleq(\overbrace{0,\ldots,0}^{L\times(u-1)},\phi(x),\overbrace{0,\ldots,0}^{L\times(|\US|-u)},\\
\mu(u|x,\theta) & =\frac{e^{\theta'\xi(x,u)}}{\sum_{u'\in\US}e^{\theta'\xi(x,u')}}.\end{align*}

\cite{BhaSutGhaLee08b} reported simulation results for two \textsc{garnet}
problems: \textsc{garnet}$(30,4,2,0.1)$ and \textsc{garnet}$(100,10,3,0.1)$.
For the \textsc{garnet}$(30,4,2,0.1)$ problem, \cite{BhaSutGhaLee08b}
used critic steps $\gamma_{n}^{c,w}$ and $\gamma_{n}^{c,\eta}$,
and actor steps $\gamma_{n}^{a}$, where\[
\gamma_{n}^{c,w}=\frac{100}{1000+n^{2/3}},\quad\gamma_{n}^{c,\eta}=0.95\gamma_{n}^{c,w},\quad\gamma_{n}^{a,\eta}=\frac{1000}{100000+n},\]
and for \textsc{garnet}$(100,10,3,0.1)$ the steps were

\[
\gamma_{n}^{c,w}=\frac{10^{5}}{10^{6}+n^{2/3}},\quad\gamma_{n}^{c,\eta}=0.95\gamma_{n}^{c,w},\quad\gamma_{n}^{a,\eta}=\frac{10^{6}}{10^{8}+n}.\]
In our simulations we used a single time scale, $\gamma_{n},$ which
was equal to $\gamma_{n}^{c,w}$ as used by \cite{BhaSutGhaLee08b}.
The basis parameters for \textsc{garnet}$(30,4,2,0.1)$ were $L=8$
and $l=3$, where for \textsc{garnet}$(100,10,3,0.1)$ they were $L=20$
and $l=5$.

\begin{figure}
\begin{centering}
\includegraphics[scale=0.48]{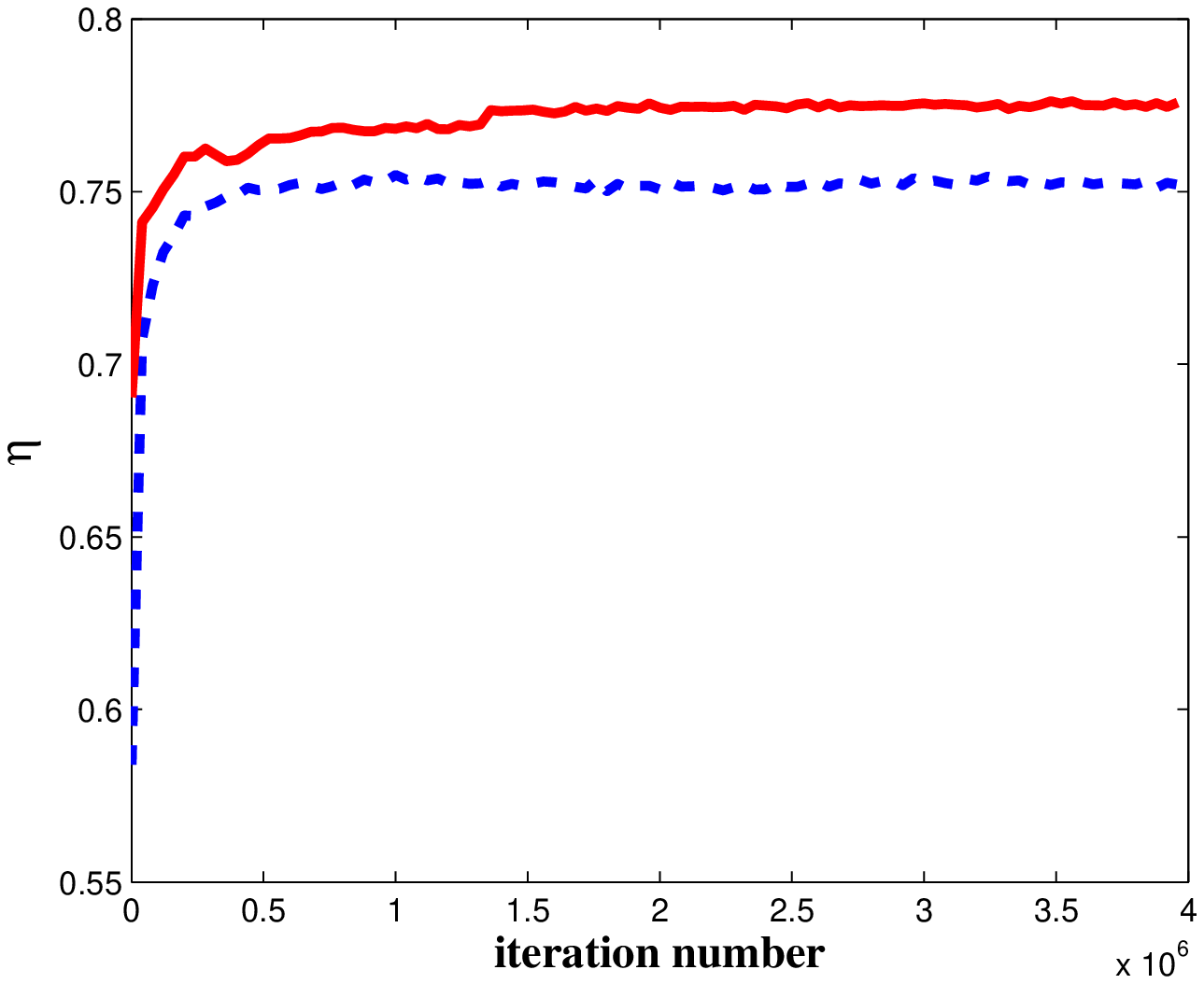}\includegraphics[scale=0.5]{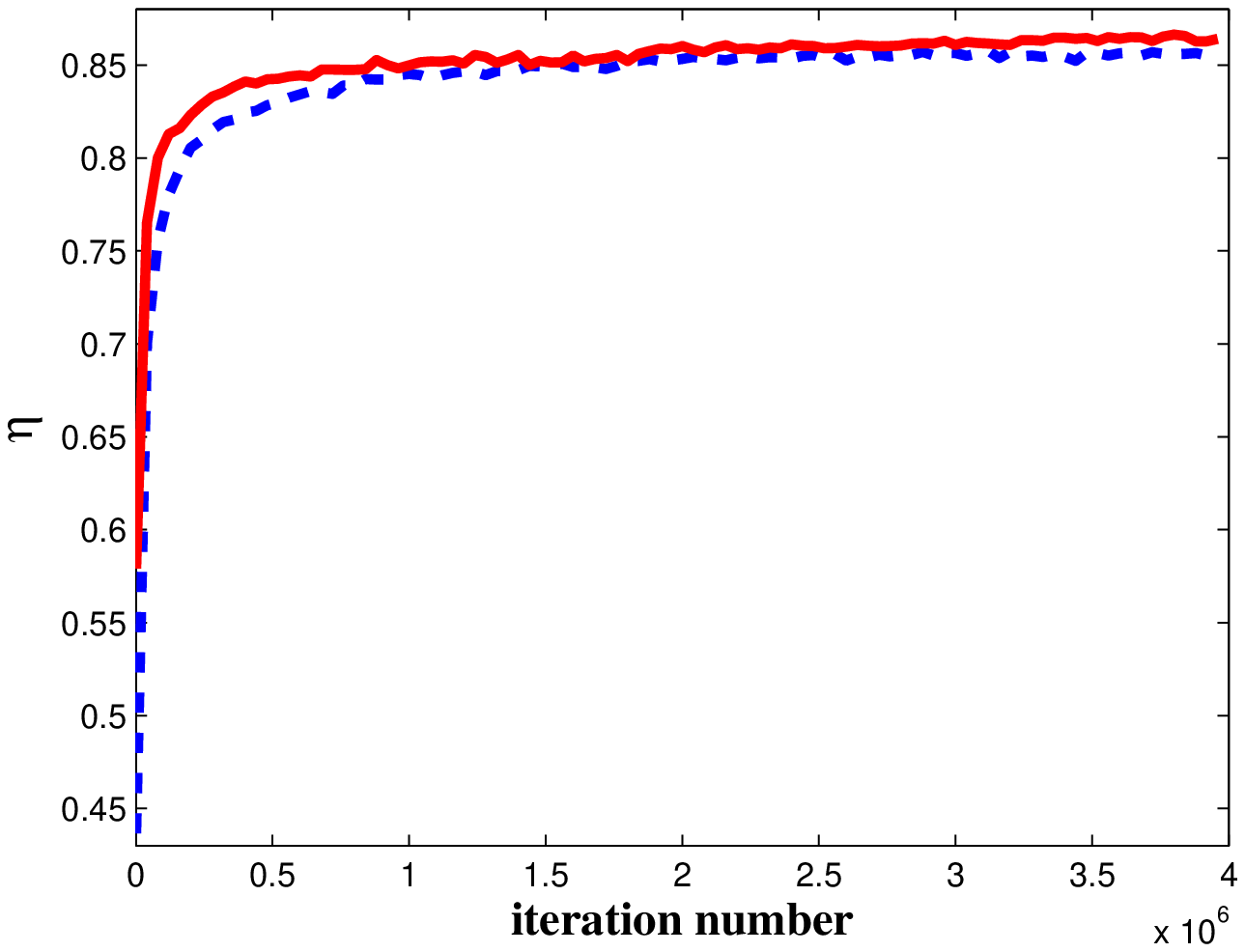}
\par\end{centering}

\begin{centering}
(a) ~~~~~~~~~~~~~~~~~~~~~~~~~~~~~~~~~~~~~~~~~~~~~~~~~~~~~~~~~~~~~~~~(b)
\par\end{centering}

\caption{Simulation results applying Algorithm 1 (red solid line) and algorithm
1 of \cite{BhaSutGhaLee08b} (blue dashed line) on a \textsc{garnet}$(30,4,2,0.1)$
problem (a) and on \textsc{garnet}$(100,10,3,0.1)$ problem (b). Standard
errors of the mean (suppressed for visibility) are of the order of
0.04.\label{fig:sim_res}}

\end{figure}

\begin{figure}
\begin{centering}
\includegraphics[scale=0.5]{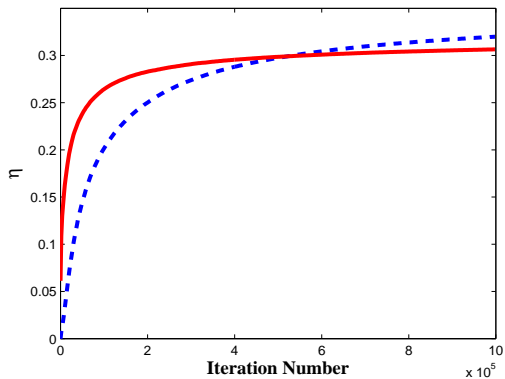}
\par\end{centering}

\caption{Simulation results applying Algorithm 1 (red solid line) and algorithm
1 of \cite{BhaSutGhaLee08b} (blue dashed line) on a \textsc{garnet}$(30,15,15,0.1)$
problem. Standard errors of the mean (suppressed for visibility) are
of the order of 0.018.\textcolor{magenta}{\label{fig:sim_counter}}}

\end{figure}

In Figures \ref{fig:sim_res} we show results of applying Algorithm
1 (solid line) and algorithm 1 from \cite{BhaSutGhaLee08b} (dashed
line) on \textsc{garnet}$(30,4,2,0.1)$ and \textsc{garnet}$(100,10,3,0.1)$
problems. Each graph in Figure \ref{fig:sim_res}, represents an average
of $100$ independent simulations. Note that an agent with a uniform
action selection policy will attain an average reward per stage of
zero in these problems. Figure \ref{fig:sim_counter} presents similar
results for \textsc{garnet}$(30,15,15,0.1)$. We see from these results
that in all simulations, during the initial phase, Algorithm \ref{algo:TD1ACFunction}
converges faster than algorithm 1 from \cite{BhaSutGhaLee08b}. The
long term behavior is problem-dependent, as can be seen by comparing
figures \ref{fig:sim_res} and \ref{fig:sim_counter}; specifically,
in Figure \ref{fig:sim_res} the present algorithm converges to a
higher value than \cite{BhaSutGhaLee08b}, while the situation is
reversed in Figure \ref{fig:sim_counter}. We refer the reader to\cite{MokPel06}
for careful discussion of convergence rates for two time scales algorithms;
a corresponding analysis of convergence rates for single time scale
algorithms is currently an open problem.

The results displayed here suggest a possible avenue for combining
both algorithms. More concretely, using the present approach may lead
to faster initial convergence due to the single time scale setting,
which allows both the actor and the critic to evolve rapidly, while
switching smoothly to a two time scales approach as in (\cite{BhaSutGhaLee08b})
will lead to asymptotic convergence to a point rather than to a region.
This type of approach is reminiscent of the quasi-Newton algorithms
in optimization, and is left for future work. As discussed in Section
\ref{sec:comparison}, we do not consider the natural gradient based
algorithms from \cite{BhaSutGhaLee08b} in this comparative study.

\section{Discussion and Future Work \label{sec:Discussion}}

We have introduced an algorithm where the information passed from
the critic to the actor is the temporal difference signal, while the
critic applies a $\textrm{TD}(\lambda)$ procedure. A policy gradient
approach was used in order to update the actor's parameters, based
on a critic using linear function approximation. The main contribution
of this work is a convergence proof in a situation where both the
actor and the critic operate on the same time scale. The drawback
of the extra flexibility in time scales is that convergence is only
guaranteed to a neighborhood of a local maximum value of the average
reward per stage. However, this neighborhood depends on parameters
which may be controlled to improve convergence.

This work sets the stage for much future work. First, as observed
above, the size of the convergence neighborhood is inversely proportional
to the step sizes $\Gamma_{w}$ and $\Gamma_{\eta}$. In other words,
in order to reduce this neighborhood we need to select larger values
of $\Gamma_{w}$ and $\Gamma_{\eta}$. This on the other hand increases
the variance of the algorithm. Therefore, further investigation of
methods which reduce this variance are needed. However, the bounds
used throughout are clearly rather loose, and cannot be effectively
used in practical applications. Obviously, improving the bounds, and
conducting careful numerical simulations in order to obtain a better
practical understanding of the influence of the different algorithmic
parameters, is called for. In addition, there is clearly room for
combining the advantages of our approach with those of AC algorithms
for which convergence to a single point is guaranteed, as discussed
in Section \ref{sec:simulations},

From a biological point of view, our initial motivation to investigate
TD based AC algorithms stemmed from questions related to the implementation
of RL in the mammalian brain. Such a view is based on an interpretation
of the transient activity of the neuromodulator dopamine as a TD signal
(e.g., \cite{Schultz02}). Recent evidence suggested that the dorsal
and ventral striatum may implement the actor and the critic, respectively
(e.g., \cite{NivDawDayan2006}). We believe that theoretical models
such as (\cite{BhaSutGhaLee08b}) and Algorithm \ref{algo:TD1ACFunction}
may provide, even if partially, a firm foundation to theories at the
neural level. Some initial attempts in a neural setting (using direct
policy gradient rather than AC based approaches) have been made by
\cite{BarasMeir2007} and \cite{Flo2007}. Such an approach may lead
to functional insights as to how an AC paradigm may be implemented
at the cellular level of the basal ganglia and cortex. An initial
demonstration was given by \cite{DICasVolMei08}.

From a theoretical perspective many issues remain open. First, strengthening
Theorem \eqref{thrm:main2_ODE_convergence} by replacing $\liminf$
by $\lim$ would clearly be useful. Second, extending the recent convergence
rate results in \cite{MokPel06} to the single time scale case is
an important challenging problem. Third, systematically combining
the advantages of single time scale convergence (fast initial dynamics)
and two time scales approaches (convergence to a point) would clearly
be beneficial.

\bigskip{}
\textbf{Acknowledgment} The authors are grateful to Mohammad Ghavamzadeh
for sending them a copy of \cite{BhaSutGhaLee08b} prior to publication,
and to the anonymous reviewers for their helpful comments. The work
of R. Meir was partially supported by an ISF Converging Technologies
grant, and by ISF grant 665/08.

\appendix
\begin{center}
\textbf{\LARGE APPENDIX}\textbf{ }
\par\end{center}

\section{Proofs of Results from Section \ref{sec:ProblemSetup}}

\subsection{Proof of Lemma \ref{lemma:pi_eta_P_bounded} \label{app:pi_eta_P_bounded}}

\begin{enumerate}
\item Looking at \eqref{eq:trans_mat_policy} we see that $P(y|x,\theta)$
is a compound function of an integral and a twice differentiable function,
$\mu(y|x,\theta)$, with bounded first and second derivatives according
to Assumption \ref{asum:mu_bound}. Therefore, $P(y|x,\theta)$ is
a twice differentiable function with bounded first and second derivatives
for all $\theta\in\R^{K}$.
\item According to Lemma \ref{lemma:ergodicy}, for each $\theta\in\R^{K}$
we have a unique solution to the following non-homogeneous linear
equation system in $\{\pi(i|\theta)\}_{i=1}^{|\XS|}$, \begin{equation}
\left\{ \begin{split}\sum_{i=1}^{|\XS|}\pi(i|\theta)P(j|i,\theta) & =\pi(j|\theta),\quad j=1,\ldots,|\XS|-1,\\
\sum_{i=1}^{|\XS|}\pi(i|\theta) & =1,\end{split}
\right.\label{eq:pi_system}\end{equation}
or in matrix form $M(\theta)\pi(\theta)=b$. By Assumption \ref{asum:aperiodicy_reccurent},
the equation system (\ref{eq:pi_system}) is invertible, therefore,
$\det[M(\theta)]>0$. This holds for all $P(\theta)\in\bar{P}$, thus,
there exists a positive constant, $b_{M}$, which uniformly lower
bounds $\det[M(\theta)]$ for all $\theta\in\R^{K}$.Thus, using Cramer's
rule we have \[
\pi(i|\theta)=\frac{Q(i,\theta)}{\det[M(\theta)]},\]
where $Q(i,\theta)$ is a finite polynomial of $\{P(j|i,\theta)\}_{i,j\in\XS}$
of at most degree $\left|\XS\right|$ and with at most $\left|\XS\right|!$
terms. Writing $\partial\pi(x|\theta)/\partial\theta_{i}$ explicitly
gives\begin{eqnarray*}
\left|\frac{\partial\pi(x|\theta)}{\partial\theta_{i}}\right| & = & \left|\frac{\det[M(\theta)]\frac{\partial}{\partial\theta_{i}}Q(i,\theta)-Q(i,\theta)\frac{\partial}{\partial\theta_{i}}\det[M(\theta)]}{\det[M(\theta)]^{2}}\right|\\
 & \le & \left|\frac{\frac{\partial}{\partial\theta_{i}}Q(i,\theta)}{\det[M(\theta)]}\right|+\left|\frac{Q(i,\theta)\frac{\partial}{\partial\theta_{i}}\det[M(\theta)]}{\det[M(\theta)]^{2}}\right|\\
 & \le & \frac{\left|\XS\right|\cdot\left|\XS\right|!\cdot B_{P_{1}}}{b_{M}}+\frac{\left(\left|\XS\right|\cdot\left|\XS\right|!\right)\cdot B_{P_{1}}}{b_{M}^{2}},\end{eqnarray*}
which gives the desired bound. Following similar steps we can show
the boundedness of the second derivatives.
\item The average reward per stage, $\eta(\theta)$ is a linear combination
of $\{\pi(i|\theta)\}_{i=1}^{|\XS|}$, with bounded coefficients by
assumption \ref{asum:r_bounded}. Therefore, using section 2, $\eta(\theta)$
is twice differentiable with bounded first and second derivatives
for all $\theta\in\R^{K}$.
\item Since $\pi(x|\theta)$ is the stationary distribution of a recurrent
MC, according to Assumption \ref{asum:aperiodicy_reccurent} there
is a positive probability to be in each state $x\in\XS$. This applies
to the closure of $\PS$. Thus, there exist a positive constant $b_{\pi}$
such that $\pi(x|\theta)\ge b_{\pi}$.
\end{enumerate}

\subsection{Proof of Lemma \ref{lemma:h_bounded} \label{app:h_bounded}}
\begin{enumerate}
\item We recall the Poisson equation \eqref{eq:Poisson_eq}. We have the
following system of linear equations in $\{h(x|\theta)\}_{x\in\XS}$,
namely, \begin{equation}
\left\{ \begin{split}h(x|\theta) & =r(x)-\eta(\theta)+\sum_{y\in\XS}P(y|x,\theta)h(y|\theta),\quad\forall x\in\XS,x\ne x^{*},\\
h(x^{*}|\theta) & =0.\end{split}
\right.\label{eq:PoissonEqn2}\end{equation}
 or in matrix form $N(\theta)h(\theta)=c$. Adding the equation $h(x^{*}|\theta)=0$
yields a unique solution for the system (\cite{Ber2006}, Vol. 1,
Prop. 7.4.1). Thus, using Cramer's rule we have $h(x|\theta)=R(x,\theta)/\det[N(\theta)]$,
where $R(x,\theta)$ and $\det[N(\theta)]$ are polynomial function
of entries in $N(\theta)$, which are bounded and have bounded first
and second derivatives according to Lemma \ref{lemma:pi_eta_P_bounded}.
Continuing in the same steps of Lemma \ref{lemma:pi_eta_P_bounded}
proof, we conclude that $h(x|\theta)$ and its two first derivatives
for all $x\in\XS$ and for all $\theta\in\R^{K}$.
\item Trivially, by (\ref{eq:TD_def}) and the previous section the result
follows.
\end{enumerate}

\section{Proof of Theorem \ref{thrm:grad_eta_based TD} \label{app:grad_eta_based_TD}}

We begin with a Lemma which was proved in (\cite{MarTsi98}). It relates
the gradient of the average reward per stage to the differential value
function.
\begin{lemma}
\label{lemma:grad_eta_based_h} The gradient of the average reward
per stage can be expressed by \begin{equation}
\nt\eta(\theta)=\sum_{x,y\in\mathcal{X},u\in\mathcal{U}}P(x,u,y,\theta)\psi(x,u,\theta)h(y,\theta).\label{eq:grad_eta}\end{equation}

\end{lemma}
For completeness, we present a proof,which will be used in the sequel.
\begin{proof}
We begin with Poisson's equation \eqref{eq:Poisson_eq} in vector
form \[
h(\theta)=\bar{r}-e\eta(\theta)+P(\theta)h(\theta),\]
 where $e$ is a column vector of $1$'s. Taking the derivative with
respect to $\theta$ and rearranging yields \[
e\nt\eta(\theta)=-\nt h(\theta)+\nt P(\theta)h(\theta)+P(\theta)\nt h(\theta).\]
 Multiplying the left hand side of the last equation by the stationary
distribution $\pi(\theta)'$ yields \[
\begin{split}\nt\eta(\theta) & =-\pi(\theta)'\nt h(\theta)+\pi\left(\theta\right)'\nt P(\theta)h(\theta)+\pi\left(\theta\right)'P(\theta)\nt h(\theta)\\
 & =-\pi\left(\theta\right)'\nt h(\theta)+\pi\left(\theta\right)'\nt P(\theta)h(\theta)+\pi\left(\theta\right)'\nt h(\theta)\\
 & =\pi\left(\theta\right)'\nt P(\theta)h(\theta).\end{split}
\]
 Expressing the result explicitly we obtain \begin{equation}
\begin{split}\nt\eta(\theta) & =\sum_{x,y\in\mathcal{X}}P(x)\nt P(y|x,\theta)h(y,\theta)\\
 & =\sum_{x,y\in\mathcal{X}}P(x)\nt\left(\sum_{u}\left(P(y|x,u)\mu(u|x,\theta)\right)\right)h(y,\theta)\\
 & =\sum_{x,y\in\mathcal{X}}P(x)\sum_{u}\left(P(y|x,u)\nt\mu(u|x,\theta)\right)h(y,\theta)\\
 & =\sum_{x,y\in\mathcal{X},u\in\mathcal{U}}P(y|x,u)P(x)\nt\mu(u|x,\theta)h(y,\theta)\\
 & =\sum_{x,y\in\mathcal{X},u\in\mathcal{U}}P(y|x,u)\mu(u|x,\theta)P(x)\frac{\nt\mu(u|x,\theta)}{\mu(u|x,\theta)}h(y,\theta)\\
 & =\sum_{x,y\in\mathcal{X},u\in\mathcal{U}}P(x,u,y,\theta)\psi(x,u,\theta)h(y,\theta).\end{split}
\label{eq:l1}\end{equation}

\end{proof}
Based on this, we can now prove Theorem \ref{thrm:grad_eta_based TD}.
We start with the result in \eqref{eq:l1}. \[
\begin{split}\nt\eta(\theta)= & \sum_{x,y\in\mathcal{X},u\in\mathcal{U}}P(x,u,y,\theta)\psi(x,u,\theta)h(y,\theta).\\
= & \sum_{x,y\in\mathcal{X},u\in\mathcal{U}}P(x,u,y,\theta)\psi(x,u,\theta)\left(h(y,\theta)-h(x,\theta)+\br(x)-\eta(\theta)+f(x)\right)\\
 & -\sum_{x,y\in\mathcal{X},u\in\mathcal{U}}P(x,u,y,\theta)\psi(x,u,\theta)\left(-h(x,\theta)+\br(x)-\eta(\theta)+f(x)\right)\\
= & \sum_{x,y\in\mathcal{X},u\in\mathcal{U}}P(x,u,y,\theta)\psi(x,u,\theta)\left(d(x,y,\theta)+f(x)\right)\\
 & -\sum_{x,y\in\mathcal{X},u\in\mathcal{U}}P(x,u,y,\theta)\psi(x,u,\theta)\left(-h(x,\theta)+\br(x)-\eta(\theta)+f(x)\right)\end{split}
\]
 In order to complete the proof, we show that the second term equals
$0$. We define $F(x,\theta)\triangleq-h(x|\theta)+\br(x)-\eta(\theta)+f(x)$
and obtain \[
\begin{split}\sum_{x,y\in\mathcal{X},u\in\mathcal{U}}P(x,u,y,\theta)\psi(x,u,\theta)F(x,\theta)= & \sum_{x\in\XS}\pi(x,\theta)F(x,\theta)\sum_{u\in\US,y\in\XS}\nt P(y|x,u,\theta)\\
= & 0.\end{split}
\]

\section{Proof of Theorem \ref{thrm:main1_algoritm_to_ODE}\label{app:main1_algoritm_to_ODE}}

As mentioned earlier, we use Theorem 6.1.1 of \cite{KusYin1997}.
We start by describing the setup of the theorem and the main result.
Then, we show that the required assumptions hold in our case.

\subsection{Setup, Assumptions and Theorem 6.1.1 of \cite{KusYin1997}.}

In this section we describe briefly but accurately the conditions
for Theorem 6.1.1 of \cite{KusYin1997} and state the main result.
We consider the following stochastic iteration\begin{equation}
y_{n+1}=\Pi_{H}[y_{n}+\gamma_{n}Y_{n}],\label{eq:y_n genenral iteration}\end{equation}
where $Y_{n}$ is a vector of {}``observations'' at time $n$, and
$\Pi_{H}$ is a constraint operator as defined in Definition \ref{asum:theta_w_constrained}.
Recall that $\left\{ x_{n}\right\} $ is a Markov chain. Based on
this, define $\F_{n}$ to be the $\sigma$-algebra\begin{eqnarray*}
\F_{n} & \triangleq & \sigma\{y_{0},Y_{i-1},x_{i}\left|i\le n\right.\}\\
 & = & \sigma\{y_{0},Y_{i-1},x_{i},y_{i}\left|i\le n\right.\},\end{eqnarray*}
and \[
\FF_{n}\triangleq\sigma\{y_{0},Y_{i-1},y_{i}\left|i\le n\right.\}.\]
The difference between the $\sigma$-algebras is the sequence $\left\{ x_{n}\right\} $.
Define the conditioned average iterate\begin{eqnarray*}
g_{n}\left(y_{n},x_{n}\right) & \triangleq & \E\left[Y_{n}\left|\F_{n}\right.\right],\end{eqnarray*}
and the corresponding \emph{martingale difference noise}\[
\delta M_{n}\triangleq Y_{n}-\E\left[Y_{n}\left|\F_{n}\right.\right].\]
Thus, we can write the iteration as\[
y_{n+1}=y_{n}+\gamma_{n}\left(g_{n}\left(y_{n},x_{n}\right)+\delta M_{n}+Z_{n}\right),\]
where $Z_{n}$ is a reflection term which forces the iterate to the
nearest point in the  set $H$ whenever the iterates leaves it (see
\cite{KusYin1997} for details). Next, set \[
\bar{g}\left(y\right)\triangleq\E\left[g_{n}\left(y,x_{n}\right)\left|\FF_{n}\right.\right].\]
Later, we will see that the sum of the sequence $\left\{ \delta M_{n}\right\} $
converges to $0$, and the r.h.s of the iteration behaves approximately
as a the function $\bar{g}\left(y\right)$, which yields the corresponding
ODE, i.e., \[
\dot{y}=\bar{g}\left(y\right).\]
The following ODE method will show that the asymptotic behavior of
the iteration is equal to the asymptotic behavior of the corresponding
ODE.

Define the auxiliary variable\[
t_{n}\triangleq\sum_{k=0}^{n-1}\gamma_{k,}\]
and the monotone piecewise constant auxiliary function \[
m\left(t\right)=\left\{ n\left|t_{n}\le t<t_{n+1}\right.\right\} .\]
The following assumption, taken from Section 6.1 of \cite{KusYin1997},
is required to establish the basic Theorem. An interpretation of the
assumption follows its statement.
\begin{assumption}
\label{ass:Kushner Yin Assumptions}Assume that \end{assumption}
\begin{enumerate}
\item The coefficients $\left\{ \gamma_{n}\right\} $ satisfy $\sum_{n=1}^{\infty}\gamma_{n}=\infty$
and $\lim_{n\rightarrow\infty}\gamma_{n}=0$.

\begin{enumerate}
\item $\sup_{n}\E\left[\left\Vert Y_{n}\right\Vert \right]<\infty.$
\item $g_{n}\left(y_{n},x\right)$ is continuous in $y_{n}$ for each $x$
and $n$.
\item For each $\mu>0$ and for some $T>0$ there is a continuous function
$\bar{g}\left(\cdot\right)$ such that for each $y$\[
\lim_{n\rightarrow\infty}\Pr\left(\sup_{j\ge n}\max_{0\le t\le T}\left\Vert \sum_{i=m\left(jT\right)}^{m\left(jT+t\right)-1}\gamma_{i}\left(g_{n}\left(y,x_{i}\right)-\bar{g}\left(y\right)\right)\right\Vert \ge\mu\right)=0.\]

\item For each $\mu>0$ and for some $T>0$ we have \[
\lim_{n\rightarrow\infty}\Pr\left(\sup_{j\ge n}\max_{0\le t\le T}\left\Vert \sum_{i=m\left(jT\right)}^{m\left(jT+t\right)-1}\gamma_{i}\delta M_{i}\right\Vert \ge\mu\right)=0.\]

\item There are measurable and non-negative functions $\rho_{3}\left(y\right)$
and $\rho_{n4}\left(x\right)$such that \[
\left\Vert g_{n}\left(y_{n},x\right)\right\Vert \le\rho_{3}\left(y\right)\rho_{n4}\left(x\right)\]
where $\rho_{3}\left(y\right)$ is bounded on each bounded $y$-set
, and for each $\mu>0$ we have\[
\lim_{\tau\rightarrow0}\lim_{n\rightarrow\infty}\Pr\left(\sup_{j\ge n}\sum_{i=m\left(j\tau\right)}^{m\left(j\tau+\tau\right)-1}\gamma_{i}\rho_{n4}\left(x_{i}\right)\ge\mu\right)=0.\]

\item There are measurable and non-negative functions $\rho_{1}\left(y\right)$
and $\rho_{n2}\left(x\right)$such that $\rho_{1}\left(y\right)$
is bounded on each bounded $y$-set and \[
\left\Vert g_{n}\left(y_{1},x\right)-g_{n}\left(y_{2},x\right)\right\Vert \le\rho_{1}\left(y_{1}-y_{2}\right)\rho_{n2}\left(x\right),\]
where\[
\lim_{y\rightarrow0}\rho_{1}\left(y\right)=0,\]
and \[
\Pr\left(\limsup_{j}\sum_{i=j}^{m\left(t_{j}+\tau\right)}\gamma_{i}\rho_{i2}\left(x_{i}\right)<\infty\right)=1.\]

\end{enumerate}
\end{enumerate}
The conditions of Assumption \ref{ass:Kushner Yin Assumptions} are
quite general but can be interpreted as follows. Assumptions \ref{ass:Kushner Yin Assumptions}.1-3
are straightforward. Assumption \ref{ass:Kushner Yin Assumptions}.4
is reminiscent of ergodicity, which is used to replace the state-dependent
function $g_{n}\left(\cdot,\cdot\right)$ with the state-independent
of state function $\bar{g}\left(\cdot\right)$, whereas Assumption
\ref{ass:Kushner Yin Assumptions}.5 states that the martingale difference
noise converges to $0$ in probability. Assumptions \ref{ass:Kushner Yin Assumptions}.6
and \ref{ass:Kushner Yin Assumptions}.7 ensure that the function
$g_{n}\left(\cdot,\cdot\right)$ is not unbounded and satisfies a
Lipschitz condition.

The following Theorem, adapted from \cite{KusYin1997}, provides the
main convergence result required. The remainder of this appendix shows
that the required conditions in Assumption \ref{ass:Kushner Yin Assumptions}
hold.
\begin{theorem}
\label{thrm:KushnerYin_project}(Adapted from Theorem 6.1.1 in \cite{KusYin1997})
Assume that algorithm \ref{algo:TD1ACFunction}, and Assumption \ref{ass:Kushner Yin Assumptions}
hold. Then $y_{n}$ converges to some invariant set of the projected
ODE \[
\dot{y}=\Pi_{H}[\bar{g}(y)].\]

\end{theorem}
Thus, the remainder of this section is devoted to showing that Assumptions
\ref{ass:Kushner Yin Assumptions}.1-\ref{ass:Kushner Yin Assumptions}.7
are satisfied.

For future purposes, we express Algorithm \ref{algo:TD1ACFunction}
using the augmented parameter vector $y_{n}$ \begin{equation}
y_{n}\triangleq\left(\theta'_{n}\quad w'_{n}\quad\EtaE'_{n}\right)',\quad\theta_{n}\in\R^{K},\quad w_{n}\in\R^{L},\quad\EtaE_{n}\in\R.\label{eq:small_y}\end{equation}
 The components of $Y_{n}$ are determined according to \eqref{eq:ODE_w}.
The corresponding sub-vectors of $\bar{g}(y_{n})$ will be denoted
by \[
\bar{g}\left(y_{n}\right)=\left[\bar{g}\left(\theta_{n}\right)'\quad\bar{g}\left(w_{n}\right)'\quad\bar{g}\left(\EtaE_{n}\right)'\right]^{'}\in\R^{K+L+1},\]
and similarly \[
g_{n}\left(y_{n},x_{n}\right)=\left[g_{n}\left(\theta_{n},x_{n}\right)'\quad g_{n}\left(w_{n},x_{n}\right)'\quad g_{n}\left(\EtaE_{n},x_{n}\right)'\right]^{'}\in\R^{K+L+1}.\]
We begin by examining the components of $g_{n}\left(y_{n},x_{n}\right)$
and $\bar{g}\left(y_{n}\right)$. The iterate $g_{n}\left(\EtaE_{n},x_{n}\right)$
is \begin{eqnarray}
g_{n}\left(\EtaE_{n},x_{n}\right) & = & \E\left[\left.\Gamma_{\eta}\left(r\left(x_{n}\right)-\ee_{n}\right)\right|\F_{n}\right]\label{eq:g_n_eta}\\
 & = & \Gamma_{\eta}\left(r\left(x_{n}\right)-\ee_{n}\right),\nonumber \end{eqnarray}
and since there is no dependence on $x_{n}$ we have also\begin{eqnarray*}
\bar{g}\left(\ee_{n}\right) & = & \Gamma_{\eta}\left(\eta\left(\theta\right)-\ee_{n}\right).\end{eqnarray*}
The iterate $g_{n}\left(w_{n},x_{n}\right)$ is \begin{eqnarray}
g_{n}\left(w_{n},x_{n}\right) & = & \E\left[\left.\Gamma_{w}\de\left(x_{n},x_{n+1},w_{n}\right)e_{n}\right|\F_{n}\right]\nonumber \\
 & = & \E\left[\left.\Gamma_{w}\sum_{k=0}^{\infty}\lambda^{k}\phi\left(x_{n-k}\right)\left(r\left(x_{n}\right)-\ee_{n}+\phi\left(x_{n+1}\right)'w_{n}-\phi\left(x_{n}\right)'w_{n}\right)\right|\F_{n}\right]\label{eq:g_n_w}\\
 & = & \Gamma_{w}\sum_{k=0}^{\infty}\lambda^{k}\phi\left(x_{n-k}\right)\left(r\left(x_{n}\right)-\ee_{n}+\sum_{y\in\XS}P\left(y|x_{n},\theta_{n}\right)\phi\left(y\right)'w_{n}-\phi\left(x_{n}\right)'w_{n}\right),\nonumber \end{eqnarray}
and the iterate $\bar{g}\left(w_{n}\right)$ is \begin{eqnarray*}
\bar{g}\left(w_{n}\right) & = & \E\left[\left.g_{n}\left(w_{n},x_{n}\right)\right|\FF_{n}\right]\\
 & = & \E\left[\left.\Gamma_{w}\sum_{k=0}^{\infty}\lambda^{k}\phi\left(x_{n-k}\right)\left(r\left(x_{n}\right)-\ee_{n}+\sum_{y\in\XS}P\left(y|x_{n},\theta_{n}\right)\phi\left(y\right)'w_{n}-\phi\left(x_{n}\right)'w_{n}\right)\right|\FF\right]\\
 & = & \Gamma_{w}\sum_{k=0}^{\infty}\lambda^{k}\sum_{x\in\XS}\pi\left(x\right)\phi\left(x\right)\sum_{z\in\XS}\left[P^{k}\right]_{xz}\left(r\left(z\right)-\ee_{n}+\sum_{y\in\XS}P\left(y|z,\theta_{n}\right)\phi\left(z\right)'w_{n}-\phi\left(y\right)'w_{n}\right),\end{eqnarray*}
which, following, \cite{BerTsi96} section 6.3, can be written in
matrix form\[
\bar{g}\left(w_{n}\right)=\Phi'\Pi\left(\theta_{n}\right)\left(\left(1-\lambda\right)\sum_{k=0}^{\infty}\lambda^{k}P^{k+1}-I\right)\Phi w_{n}+\Phi'\Pi\left(\theta_{n}\right)\sum_{k=0}^{\infty}\lambda^{k}P^{k}\left(r-\ee_{n}\right).\]
With some further algebra we can express this using (\ref{eq:thrm_main_vars_w}),
\[
\bar{g}\left(w_{n}\right)=A\left(\theta_{n}\right)w_{n}+b\left(\theta_{n}\right)+G\left(\theta_{n}\right)\left(\eta\left(\theta_{n}\right)-\ee_{n}\right).\]
Finally, the iterate $g_{n}\left(\theta_{n},x_{n}\right)$ is \begin{eqnarray}
g_{n}\left(\theta_{n},x_{n}\right) & = & \E\left[\left.\de\left(x_{n},x_{n+1},w_{n}\right)\psi\left(x_{n},u_{n},\theta_{n}\right)\right|\F_{n}\right]\nonumber \\
 & = & \E\left[\left.d\left(x_{n},x_{n+1},\theta_{n}\right)\psi\left(x_{n},u_{n},\theta_{n}\right)\right|\F_{n}\right]\label{eq:g_n_theta}\\
 &  & +\E\left[\left.\left(\de\left(x_{n},x_{n+1},w_{n}\right)-d\left(x_{n},x_{n+1},\theta_{n}\right)\right)\psi\left(x_{n},u_{n},\theta_{n}\right)\right|\F_{n}\right]\nonumber \\
 & = & \E\left[\left.d\left(x_{n},x_{n+1},\theta_{n}\right)\psi\left(x_{n},u_{n},\theta_{n}\right)\right|\F_{n}\right]\nonumber \\
 &  & +\sum_{z\in\XS}P\left(z|x_{n}\right)\psi\left(x_{n},u_{n},\theta_{n}\right)\left(\de\left(x_{n},z,w_{n}\right)-d\left(x_{n},z,\theta_{n}\right)\right),\nonumber \end{eqnarray}
and \begin{eqnarray*}
\bar{g}\left(\theta_{n}\right) & = & \E\left[\left.\de\left(x_{n},x_{n+1},w_{n}\right)\psi\left(x_{n},u_{n},\theta_{n}\right)\right|\FF_{n}\right]\\
 & = & \E\left[\left.d\left(x_{n},x_{n+1},\theta_{n}\right)\psi\left(x_{n},u_{n},\theta_{n}\right)\right|\FF_{n}\right]+\E\left[\left.\left(\de\left(x_{n},x_{n+1},w_{n}\right)-d\left(x_{n},x_{n+1},\theta_{n}\right)\right)\psi\left(x_{n},u_{n},\theta_{n}\right)\right|\FF_{n}\right]\\
 & = & \nabla\eta(\theta_{n})+\sum_{x,y\in\XS}\sum_{u\in\US}\pi\left(x\right)P\left(u|x,\theta_{n}\right)P\left(y|x,u\right)\psi\left(x,u,\theta_{n}\right)\left(\de\left(x,y,w_{n}\right)-d\left(x,y,\theta_{n}\right)\right).\end{eqnarray*}

Next, we show that the required assumptions hold.

\subsection{Satisfying Assumption \ref{ass:Kushner Yin Assumptions}.2}

We need to show that $\sup_{n}\E\left[\left\Vert Y_{n}\right\Vert _{2}\right]<\infty.$
Since later we need to show that $\sup_{n}\E\left[\left\Vert Y_{n}\right\Vert _{2}^{2}\right]<\infty$,
and the proof of the second moment is similar to the proof of the
first moment, we consider both moments here.
\begin{lemma}
\label{lemma:bounded_eta_est} The sequence $\EtaE_{n}$ is bounded
w.p. $\!1$, $\sup_{n}\mathrm{\E}\left[\left\Vert Y_{n}\left(\ee_{n}\right)\right\Vert _{2}\right]<\infty$,
and $\sup_{n}\mathrm{\E}\left[\left\Vert Y_{n}\left(\ee_{n}\right)\right\Vert _{2}^{2}\right]<\infty$\end{lemma}
\begin{proof}
We can choose $M$ such that $\gamma_{n}\Gamma_{\eta}<1$ for all
$n>M$. Using Assumption \ref{asum:mu_bound} for the boundedness
of the rewards, we have \begin{equation}
\begin{split}\EtaE_{n+1} & =(1-\gamma_{n}\Gamma_{\eta})\EtaE_{m}+\gamma_{n}\Gamma_{\eta}r(x_{n})\\
 & \le(1-\gamma_{n}\Gamma_{\eta})\EtaE_{n}+\gamma_{n}\Gamma_{\eta}B_{r}\\
 & \le\left\{ \begin{array}{lcl}
\EtaE_{n} &  & \textrm{if }\EtaE_{n}>B_{r},\\
B_{r} &  & \textrm{if }\EtaE_{n}\le B_{r},\end{array}\right.\\
 & \le\max\{\EtaE_{n},B_{r}\},\end{split}
\label{eq:eta_is_bounded}\end{equation}
 which means that each iterate is bounded above by the previous iterate
or by a constant. We denote this bound by $B_{\tilde{\eta}}$. Using
similar arguments we can prove that $\EtaE_{n}$ is bounded below,
and the first part of the lemma is proved. Since $\EtaE_{n+1}$ is
bounded the second part follows trivially.\end{proof}
\begin{lemma}
\label{lemma:bounded_w} We have $\sup_{n}\E\left[\left\Vert Y_{n}\left(w_{n}\right)\right\Vert _{2}^{2}\right]<\infty$
and $\sup_{n}\E\left[\left\Vert Y_{n}\left(w_{n}\right)\right\Vert _{2}\right]<\infty$\end{lemma}
\begin{proof}
For the first part we have\begin{eqnarray*}
\E\left[\left\Vert Y_{n}\left(w_{n}\right)\right\Vert _{2}^{2}\right] & = & \E\left[\left\Vert \Gamma_{w}\de\left(x_{n},x_{n+1},w_{n}\right)e_{n}\right\Vert _{2}^{2}\right]\\
 & = & \Gamma_{w}^{2}\E\left[\left\Vert \sum_{k=0}^{\infty}\lambda^{k}\phi\left(x_{n-k}\right)\left(r\left(x_{n}\right)-\ee_{n}+\phi\left(x_{n+1}\right)'w_{n}-\phi\left(x_{n}\right)'w_{n}\right)\right\Vert _{2}^{2}\right]\\
 & \stackrel{(a)}{\le} & \Gamma_{w}^{2}\E\left[\sum_{k=0}^{\infty}\lambda^{k}\left\Vert \phi\left(x_{n-k}\right)\left(r\left(x_{n}\right)-\ee_{n}+\phi\left(x_{n+1}\right)'w_{n}-\phi\left(x_{n}\right)'w_{n}\right)\right\Vert _{2}\right]^{2}\\
 & \le & \Gamma_{w}^{2}\E\left[\sup_{k}\left\Vert \phi\left(x_{n-k}\right)\left(r\left(x_{n}\right)-\ee_{n}+\phi\left(x_{n+1}\right)'w_{n}-\phi\left(x_{n}\right)'w_{n}\right)\right\Vert _{2}\sum_{k=0}^{\infty}\lambda^{k}\right]^{2}\\
 & \stackrel{(b)}{\le} & \frac{4\Gamma_{w}^{2}}{\left(1-\lambda\right)^{2}}\left\Vert \phi\left(x_{n-k}\right)\right\Vert _{2}^{2}\left(\left|r\left(x_{n}\right)\right|^{2}+\left|\ee_{n}\right|^{2}+\left\Vert \phi\left(x_{n+1}\right)\right\Vert _{2}^{2}\cdot\left\Vert w_{n}\right\Vert _{2}^{2}+\left\Vert \phi\left(x_{n}\right)\right\Vert _{2}^{2}\cdot\left\Vert w_{n}\right\Vert _{2}^{2}\right)\\
 & \le & \frac{4\Gamma_{w}^{2}}{\left(1-\lambda\right)^{2}}B_{\phi}^{2}\left(B_{r}^{2}+B_{\tilde{\eta}}^{2}+2B_{\phi}^{2}B_{w}^{2}\right),\end{eqnarray*}
where we used the triangle inequality in $(a)$ and the inequality
$(a+b)^{2}\le2a^{2}+2b^{2}$ in $(b)$. The bound $\sup_{n}\E\left[\left\Vert Y_{n}\left(w_{n}\right)\right\Vert _{2}\right]<\infty$
follows directly from the Cauchy-Schwartz inequality.\end{proof}
\begin{lemma}
\label{lemma:bounded_theta} We have $\sup_{n}\E\left[\left\Vert Y_{n}\left(\theta_{n}\right)\right\Vert _{2}^{2}\right]<\infty$
and $\sup_{n}\E\left[\left\Vert Y_{n}\left(\theta_{n}\right)\right\Vert _{2}\right]<\infty$.The
proof proceeds as in Lemma \ref{lemma:bounded_w}.
\end{lemma}
Based on Lemmas \ref{lemma:bounded_eta_est}, \ref{lemma:bounded_w},
and \ref{lemma:bounded_theta} we can assert Assumption \ref{ass:Kushner Yin Assumptions}.2

\subsection{Satisfying Assumption \ref{ass:Kushner Yin Assumptions}.3}

Assumption \ref{ass:Kushner Yin Assumptions}.3 requires the continuity
of $g_{n}\left(y_{n},x_{n}\right)$ for each $n$ and $x_{n}$. Again,
we show that this assumption holds for the three parts of the vector
$y_{n}$.
\begin{lemma}
The function $g_{n}\left(\EtaE_{n},x_{n}\right)$ is a continuous
function of $\ee_{n}$ for each $n$ and $x_{n}$.\end{lemma}
\begin{proof}
Since $g_{n}\left(\EtaE_{n},x_{n}\right)=\Gamma_{\eta}\left(r\left(x_{n}\right)-\ee_{n}\right)$
the claim follows. \end{proof}
\begin{lemma}
\label{lem:w_n_continuous}The function $g_{n}\left(w_{n},x_{n}\right)$
is a continuous function of $\ee_{n}$, $w_{n}$, and $\theta_{n}$
for each $n$ and $x_{n}$.\end{lemma}
\begin{proof}
The function is \[
g_{n}\left(w_{n},x_{n}\right)=\Gamma_{w}\sum_{k}^{\infty}\lambda^{k}\phi\left(x_{n-k}\right)\left(r\left(x_{n}\right)-\ee_{n}+\sum_{y\in\XS}P\left(y|x_{n},\theta_{n}\right)\phi\left(y\right)'w_{n}-\phi\left(x_{n}\right)'w_{n}\right).\]
The probability transition $\sum_{y\in\XS}P\left(y|x_{n},\theta_{n}\right)$
can be written as $\sum_{y\in\XS,u\in\US}P\left(y|x_{n},u_{n}\right)\mu\left(u_{n}|x_{n},\theta_{n}\right)$.
The function $\mu\left(u_{n}|x_{n},\theta_{n}\right)$ is continuous
in $\theta_{n}$ by Assumption \ref{lemma:pi_eta_P_bounded}, and
thus $g_{n}\left(w_{n},x_{n}\right)$ is continuous in $\ee_{n}$
and $\theta_{n}$ and the lemma follows.\end{proof}
\begin{lemma}
The function $g_{n}\left(\theta_{n},x_{n}\right)$ is a continuous
function of $\ee_{n}$, $w_{n}$, and $\theta_{n}$ for each $n$
and $x_{n}$.\end{lemma}
\begin{proof}
By definition, the function $g_{n}\left(\theta_{n},x_{n}\right)$
is\begin{eqnarray*}
g_{n}\left(\theta_{n},x_{n}\right) & = & \E\left[\left.\de\left(x_{n},x_{n+1},w_{n}\right)\psi\left(x_{n},u_{n},\theta_{n}\right)\right|\F_{n}\right]\\
 & = & \frac{\nabla_{\theta}\mu\left(u_{n}|x_{n},\theta_{n}\right)}{\mu\left(u_{n}|x_{n},\theta_{n}\right)}\left(r\left(x_{n}\right)-\ee_{n}+\sum_{y\in\XS}P\left(y|x_{n},\theta_{n}\right)\phi\left(y\right)'w_{n}-\phi\left(x_{n}\right)'w_{n}\right)\end{eqnarray*}
Using similar arguments to Lemma \ref{lem:w_n_continuous} the claim
holds.
\end{proof}

\subsection{Satisfying Assumption \ref{ass:Kushner Yin Assumptions}.4\label{sub:Sat-Assum-Kushner-yin-4}}

In this section we prove the following convergence result: for each
$\mu>0$ and for some $T>0$ there is a continuous function $\bar{g}\left(\cdot\right)$
such that for each $y$\begin{equation}
\lim_{n\rightarrow\infty}\Pr\left(\sup_{j\ge n}\max_{0\le t\le T}\left\Vert \sum_{i=m\left(jT\right)}^{m\left(jT+t\right)-1}\gamma_{i}\left(g_{n}\left(y,x_{i}\right)-\bar{g}\left(y\right)\right)\right\Vert \ge\mu\right).\label{eq:assumption_4_expression}\end{equation}
 We start by showing that there exist independent cycles of the algorithm
since the underlying Markov chain is recurrent and aperiodic. Then,
we show that the cycles behave as a martingale, thus Doob's inequality
can be used. Finally we show that the sum in (\ref{eq:assumption_4_expression})
converges to $0$ w.p. 1. We start investigating the regenerative
nature of the process.

Based on Lemma \ref{asum:aperiodicy_reccurent}, there exists a recurrent
state common to all $MC(\theta)$, denoted by $x^{*}$. We define
the series of \emph{hitting times} of the recurrent state $x^{*}$
by $t_{0}=0,t_{1},t_{2},...$, where $t_{m}$ it the $m$-th time
the agent hits the state $x^{*}$. Mathematically, we can define this
series recursively by \[
t_{m+1}=\inf\{n|x_{n}=x^{*},n>t_{m}\},\quad t_{0}=0,\]
and $T_{m}\triangleq t_{m+1}-t_{m}$. Define the $m$-th cycle of
the algorithm to be the set of times \begin{equation}
\mathcal{T}_{m}\triangleq\{n|t_{m-1}\le n<t_{m}\},\label{eq:tau_m_mth_cycle_times}\end{equation}
 and the corresponding trajectories \begin{equation}
\mathcal{C}_{m}\triangleq\{x_{n}|n\in\mathcal{T}_{m}\}.\label{eq:C_m_mth_trajectory}\end{equation}
Define a function, $\varrho\left(k\right)$, which returns the cycle
to which the time $k$ belongs to, i.e.,\[
\varrho\left(k\right)\triangleq\left\{ m\left|k\in\mathcal{T}_{m}\right.\right\} .\]
We notice that based on Lemma \ref{lemma:ergodicy}, and using the
\emph{Regenerative Cycle Theorem} (see \cite{Bremaud1999}, pp. 87),
the cycles $\mathcal{C}_{m}$ are independent of each other.

Next, we examine (\ref{eq:assumption_4_expression}), and start by
defining the following events: \begin{eqnarray*}
b_{n}^{\left(1\right)} & \triangleq & \left\{ \omega\left|\sup_{j\ge n}\max_{0\le t\le T}\left\Vert \sum_{i=m\left(jT\right)}^{m\left(jT+t\right)-1}\gamma_{i}\left(g_{i}\left(y,x_{i}\right)-\bar{g}\left(y\right)\right)\right\Vert \ge\mu\right.\right\} ,\\
b_{n}^{\left(2\right)} & \triangleq & \left\{ \omega\left|\sup_{j\ge n}\sup_{k\ge m\left(jT\right)}\left\Vert \sum_{i=m\left(jT\right)}^{k}\gamma_{i}\left(g_{i}\left(y,x_{i}\right)-\bar{g}\left(y\right)\right)\right\Vert \ge\mu\right.\right\} ,\\
b_{n}^{\left(3\right)} & \triangleq & \left\{ \omega\left|\sup_{j\ge n}\left\Vert \sum_{i=n}^{\infty}\gamma_{i}\left(g_{i}\left(y,x_{i}\right)-\bar{g}\left(y\right)\right)\right\Vert \ge\mu\right.\right\} .\end{eqnarray*}
It is easy to show that for each $n$ we have $b_{n}^{\left(1\right)}\subset b_{n}^{\left(2\right)},$
thus, \begin{equation}
\Pr\left(b_{n}^{\left(1\right)}\right)\le\Pr\left(b_{n}^{\left(2\right)}\right).\label{eq:c_bounde_by_b}\end{equation}
It is easy to verify that the series $\left\{ b_{n}^{\left(2\right)}\right\} $
is a subsequence of $\left\{ b_{n}^{\left(3\right)}\right\} $. Thus,
if we prove that $\lim_{n\rightarrow\infty}\Pr\left(b_{n}^{\left(3\right)}\right)=0$,
then $\lim_{n\rightarrow\infty}\Pr\left(b_{n}\right)=0$, and using
(\ref{eq:c_bounde_by_b}), Assumption \ref{ass:Kushner Yin Assumptions}.4
holds.

Next, we examine the sum defining the event $b_{n}^{\left(3\right)}$,
by splitting it a sum over cycles and a sum within each cycle. We
can write it as following\[
\sum_{i=n}^{\infty}\gamma_{i}\left(g_{i}\left(y,x_{i}\right)-\bar{g}\left(y\right)\right)=\sum_{m=\varrho\left(n\right)}^{\infty}\sum_{i\in\mathcal{T}_{m}}\gamma_{i}\left(g_{i}\left(y,x_{i}\right)-\bar{g}\left(y\right)\right).\]
Denote $c_{m}\triangleq\sum_{j\in\mathcal{T}_{m}}\gamma_{i}\left(g_{n}\left(y,x_{i}\right)-\bar{g}\left(y\right)\right)$.
Therefore, by the \emph{Regenerative Cycle Theorem} (\cite{Bremaud1999},
pp. 87), $c_{m}$ are independent random variables. Also,\begin{eqnarray*}
\E\left[c_{m}\right]=\E\left[\sum_{i\in\mathcal{T}_{m}}\gamma_{i}\left(g_{i}\left(y,x_{i}\right)-\bar{g}\left(y\right)\right)\right] & = & \E\left[\E\left[\left.\sum_{j\in\mathcal{T}_{m}}\gamma_{i}\left(g_{n}\left(y,x_{i}\right)-\bar{g}\left(y\right)\right)\right|\mathcal{T}_{m}\right]\right]=0.\end{eqnarray*}
We argue that $c_{m}$ is square integrable. To prove this we need
to show that the second moments of $T_{m}$ and $\left(g_{n}\left(y,x_{i}\right)-\bar{g}\left(y\right)\right)$
are finite.
\begin{lemma}
$\quad$\label{lem:c_m is square integrable}\end{lemma}
\begin{enumerate}
\item The first two moments of the random times $\left\{ T_{m}\right\} $
are bounded above by a constant $B_{T}$, for all $\theta\in\R^{K}$
and for all $m$, $1\le m<\infty.$

\begin{enumerate}
\item $\E\left[\left(g_{n}\left(y,x_{i}\right)-\bar{g}\left(y\right)\right)^{2}\right]\le B_{g}$
\item Define $\bar{\gamma}{}_{m}\triangleq\sup_{i\in\mathcal{T}_{m}}\gamma_{i}$,
then $\sum_{m=0}^{\infty}\bar{\gamma}{}_{m}^{2}<\infty$.
\item $\E\left[c_{m}^{2}\right]\le\left(B_{T}B_{g}\right)^{2}$.
\end{enumerate}
\end{enumerate}
\begin{proof}
$\quad$\end{proof}
\begin{enumerate}
\item According to Assumption \ref{asum:aperiodicy_reccurent} and Lemma
\ref{lemma:ergodicy}, each Markov chain in $\bar{\PS}$ is recurrent.
Thus, for each $\theta\in\R^{K}$ there exists a constant $\tilde{B}_{T}(\theta)$,
$0<\tilde{B}_{T}(\theta)<1$, where for $k\le\left|\XS\right|$ we
have \begin{equation}
P(T_{m}=k|\theta_{m})\le\left(\tilde{B}_{T}(\theta_{m})\right)^{\lfloor k/|\XS|\rfloor},\quad1\le m<\infty,\quad1\le k<\infty,\label{eq:bound_on_probability_Tm}\end{equation}
 where $\lfloor a\rfloor$ is the largest integer which is not greater
than $a$. Otherwise, if for $k>\left|\XS\right|$ we have $\tilde{B}_{T}(\theta_{m})=1$
then the chain transitions equal $1$ which contradicts the aperiodicity
of the chains. Therefore, \[
\E\left[\left.T_{m}\right|\theta_{m}\right]=\sum_{k=1}^{\infty}kP(T_{m}=k|\theta_{m})\le\sum_{k=1}^{\infty}k\left(\tilde{B}_{T}(\theta_{m})\right)^{\lfloor k/|\XS|\rfloor}=B_{T_{1}}(\theta_{m})<\infty,\]
 and \[
\E\left[\left.T_{m}^{2}\right|\theta_{m}\right]=\sum_{k=1}^{\infty}k^{2}P(T_{m}=k|\theta_{m})\le\sum_{k=1}^{\infty}k^{2}\left(\tilde{B}_{T}(\theta_{m})\right)^{\lfloor k/|\XS|\rfloor}=B_{T_{2}}(\theta_{m})<\infty.\]
Since the set $\bar{\PS}$ is closed, by Assumption \ref{asum:aperiodicy_reccurent}
the above holds for the closure of $\bar{\PS}$ as well. Thus, there
exists a constant $B_{T}$ satisfying $B_{T}=\max\{\sup_{\theta}B_{T_{1}}(\theta),\sup_{\theta}B_{T_{2}}(\theta)\}<\infty$.

\begin{enumerate}
\item The proof proceeds along the same lines as the proofs of lemmas \ref{lemma:bounded_eta_est},
\ref{lemma:bounded_w}, and \ref{lemma:bounded_theta}.
\item The result follows trivially since the sequence $\left\{ \bar{\gamma}{}_{m}\right\} $
is subsequence of the summable sequence$\left\{ \gamma_{m}\right\} $.
\item By definition, for large enough $m$ we have $\gamma_{m}\le1$. Therefore,
we have \begin{eqnarray*}
\E\left[c_{m}^{2}\right] & = & \E\left[\left(\sum_{j\in\mathcal{T}_{m}}\gamma_{j}\left(g_{n}\left(y,x_{j}\right)-\bar{g}\left(y\right)\right)\right)^{2}\right]\\
 & \le & \E\left[\left|\mathcal{T}_{m}\right|^{2}\left(\sup_{j}\gamma_{j}\right)^{2}\left(\sup_{j}\left(g_{n}\left(y,x_{j}\right)-\bar{g}\left(y\right)\right)\right)^{2}\right]\\
 & \le & B_{T}^{2}B_{g}^{2}.\end{eqnarray*}

\end{enumerate}
\end{enumerate}
Next, we conclude by showing that Assumption \ref{ass:Kushner Yin Assumptions}.4
is satisfied. Define the process $d_{n}\triangleq\sum_{m=0}^{n}c_{m}$.
This process is a martingale since the sequence $\left\{ c_{m}\right\} $
is square integrable (by Lemma \ref{lem:c_m is square integrable})
and satisfies $\E\left[d_{m+1}|d_{m}\right]=d_{m}$. Using Doob's
martingale inequality%
\footnote{If $w_{n}$ is a martingale sequence then $\Pr\left(\sup_{m\ge0}\left|w_{n}\right|\ge\mu\right)\le\lim_{n\rightarrow\infty}\E\left[\left|w_{n}\right|^{2}\right]/\mu^{2}$. %
} we have \begin{eqnarray*}
\Pr\left(\sup_{k\ge n}\sum_{m=\varrho\left(n\right)}^{\varrho\left(k\right)}\sum_{j\in\mathcal{T}_{m}}\gamma_{i}\left(g_{n}\left(y,x_{i}\right)-\bar{g}\left(y\right)\right)\ge\mu\right) & \le & \lim_{n\rightarrow\infty}\frac{\E\left[\left(\sum_{m=\varrho\left(n\right)}^{\infty}\sum_{j\in\mathcal{T}_{m}}\gamma_{j}\left(g_{n}\left(y,x_{j}\right)-\bar{g}\left(y\right)\right)\right)^{2}\right]}{\mu^{2}}\\
 & = & \lim_{n\rightarrow\infty}\frac{\sum_{m=\varrho\left(n\right)}^{\infty}\E\left[\left(\sum_{j\in\mathcal{T}_{m}}\gamma_{j}\left(g_{n}\left(y,x_{j}\right)-\bar{g}\left(y\right)\right)\right)^{2}\right]}{\mu^{2}}\\
 & \le & \lim_{n\rightarrow\infty}\sum_{m=\varrho\left(n\right)}^{\infty}\bar{\gamma}_{m}^{2}B_{g}B_{T}/\mu^{2}\\
 & = & 0.\end{eqnarray*}

\subsection{Satisfying Assumption \ref{ass:Kushner Yin Assumptions}.5}

In this section we need to show that for each $\mu>0$ and for some
$T>0$ we have \begin{equation}
\lim_{n\rightarrow\infty}\Pr\left(\sup_{j\ge n}\max_{0\le t\le T}\left\Vert \sum_{i=m\left(jT\right)}^{m\left(jT+t\right)-1}\gamma_{i}\delta M_{i}\right\Vert \ge\mu\right)=0.\label{eq:martingale-diff-assum}\end{equation}
In order to follow the same lines as in Section \ref{sub:Sat-Assum-Kushner-yin-4},
we need to show that the second moment of the martingale difference
noise, $\delta M_{i}$, is bounded with zero mean. By definition,
$\delta M_{n}\left(\cdot\right)$ has zero mean.
\begin{lemma}
The martingale difference noise, $\delta M_{n}\left(\cdot\right)$,
is bounded in the second moment.\label{lem:The-martingale-difference-bounded}\end{lemma}
\begin{proof}
The claim is immediate from the fact that \[
\E\left[\left(\delta M_{n}\right)^{2}\right]=\E\left[\left\Vert Y_{n}-g_{n}\left(y_{n},x_{n}\right)\right\Vert ^{2}\right]\le2\E\left[\left\Vert Y_{n}\right\Vert ^{2}+\left\Vert g_{n}\left(y_{n},x_{n}\right)\right\Vert ^{2}\right],\]
 and from Lemma \ref{lemma:bounded_eta_est}, Lemma \ref{lemma:bounded_w},
and Lemma \ref{lemma:bounded_theta}.
\end{proof}
Combining this fact with Lemma \ref{lem:The-martingale-difference-bounded},
and applying the regenerative decomposition of Section \ref{sub:Sat-Assum-Kushner-yin-4},
we conclude that statistically $\delta M_{n}\left(\cdot\right)$ behaves
exactly as $\left(g_{n}\left(y,x_{i}\right)-\bar{g}\left(y\right)\right)$
of section  \ref{sub:Sat-Assum-Kushner-yin-4} and thus (\ref{eq:martingale-diff-assum})
holds.

\subsection{Satisfying Assumption \ref{ass:Kushner Yin Assumptions}.6\label{sec:Satisfying-Assumption-.6}}

In this section we need to prove that there are non-negative measurable
functions $\rho_{3}\left(y\right)$ and $\rho_{n4}\left(x\right)$
such that \begin{equation}
\left\Vert g_{n}\left(y_{n},x\right)\right\Vert \le\rho_{3}\left(y_{n}\right)\rho_{n4}\left(x\right),\label{eq:g_n_bounded_r3rn4}\end{equation}
where $\rho_{3}\left(y\right)$ is bounded on each bounded $y$-set,
and for each $\mu>0$ we have\[
\lim_{\tau\rightarrow0}\lim_{n\rightarrow\infty}\Pr\left(\sup_{j\ge n}\sum_{i=m\left(j\tau\right)}^{m\left(j\tau+\tau\right)-1}\gamma_{i}\rho_{n4}\left(x_{i}\right)\ge\mu\right)=0.\]
The following lemma states a stronger condition for Assumption \ref{ass:Kushner Yin Assumptions}.6.
In fact, we choose $\rho_{3}(y)$ to be a positive constant.
\begin{lemma}
\label{lem:If_g_n_bounded}If \textup{$\left\Vert g_{n}\left(y,x\right)\right\Vert $
is uniformly bounded for each $y$, $x$ and $n$, then }Assumption
\ref{ass:Kushner Yin Assumptions}.6 is satisfied.\end{lemma}
\begin{proof}
Let us denote the upper bound by the random variable $B$, i.e., \[
\left\Vert g_{n}\left(y,x\right)\right\Vert \le B,\quad\textrm{w.p. 1}.\]
Thus \begin{eqnarray*}
\lim_{\tau\rightarrow0}\lim_{n\rightarrow\infty}\Pr\left(\sup_{j\ge n}\sum_{i=m\left(j\tau\right)}^{m\left(j\tau+\tau\right)-1}\gamma_{i}\rho_{n4}\left(x_{i}\right)\ge\mu\right) & \le & \lim_{\tau\rightarrow0}\lim_{n\rightarrow\infty}\Pr\left(\sup_{j\ge n}\sum_{i=m\left(j\tau\right)}^{m\left(j\tau+\tau\right)-1}\gamma_{i}B\ge\mu\right)\\
 & = & \lim_{\tau\rightarrow0}\lim_{n\rightarrow\infty}\Pr\left(\sup_{j\ge n}B\sum_{i=m\left(j\tau\right)}^{m\left(j\tau+\tau\right)-1}\gamma_{i}\ge\mu\right)\\
 & \le & \lim_{\tau\rightarrow0}\Pr\left(B\tau\ge\mu\right)\\
 & = & 0.\end{eqnarray*}

\end{proof}
Based on Lemma \ref{lem:If_g_n_bounded}, we are left with proving
that $g_{n}\left(y,x\right)$ is uniformly bounded. The following
lemma states so.
\begin{lemma}
The function \textup{$g_{n}\left(y,x\right)$ is} uniformly bounded
for all $n$.\end{lemma}
\begin{proof}
We examine the components of $g_{n}\left(y_{n},x_{n}\right)$. In
(\ref{eq:g_n_eta}) we showed that \begin{eqnarray*}
g_{n}\left(\EtaE_{n},x_{n}\right) & = & \Gamma_{\eta}\left(r\left(x_{n}\right)-\ee_{n}\right).\end{eqnarray*}
Since both $r\left(x_{n}\right)$ and $\ee_{n}$ are bounded by Assumption
\ref{asum:r_bounded} and Lemma \ref{lemma:bounded_eta_est} respectively,
we have a uniform bound on $g_{n}\left(\EtaE_{n},x_{n}\right)$. Recalling
(\ref{eq:g_n_w}) we have \begin{eqnarray*}
g_{n}\left(w_{n},x_{n}\right) & = & \Gamma_{w}\sum_{k=0}^{\infty}\lambda^{k}\phi\left(x_{n-k}\right)\left(r\left(x_{n}\right)-\ee_{n}+\sum_{y\in\XS}P\left(y|x_{n},\theta_{n}\right)\phi\left(y\right)'w_{n}-\phi\left(x_{n}\right)'w_{n}\right)\\
 & \le & \Gamma_{w}\frac{1}{1-\lambda}B_{\phi}\left(B_{r}+B_{\ee}+2B_{\phi}B_{w}\right).\end{eqnarray*}
Finally, recalling (\ref{eq:g_n_theta}) we have\begin{eqnarray*}
g_{n}\left(\theta_{n},x_{n}\right) & = & \E\left[\left.\de\left(x_{n},x_{n+1},w_{n}\right)\psi\left(x_{n},u_{n},\theta_{n}\right)\right|\F_{n}\right]\\
 & \le & \left(B_{r}+B_{\ee}+2B_{\phi}B_{w}\right)B_{\psi}.\end{eqnarray*}

\end{proof}

\subsection{Satisfying Assumption \ref{ass:Kushner Yin Assumptions}.7}

In this section we show that there are non-negative measurable functions
$\rho_{1}\left(y\right)$ and $\rho_{n2}\left(x\right)$ such that
$\rho_{1}\left(y\right)$ is bounded on each bounded $y$-set and
\begin{equation}
\left\Vert g_{n}\left(y_{1},x\right)-g_{n}\left(y_{2},x\right)\right\Vert \le\rho_{1}\left(y_{1}-y_{2}\right)\rho_{n2}\left(x\right)\label{eq:lipschitz_rho1}\end{equation}
where\begin{equation}
\lim_{y\rightarrow0}\rho_{1}\left(y\right)=0,\label{eq:rho1_to_0}\end{equation}
and for some $\tau>0$ \[
\Pr\left(\limsup_{j}\sum_{i=j}^{m\left(t_{j}+\tau\right)}\gamma_{i}\rho_{i2}\left(x_{i}\right)<\infty\right)=1.\]
From Section \ref{sec:Satisfying-Assumption-.6} we infer that we
can choose $\rho_{n2}\left(x\right)$ to be a constant since $g_{n}\left(y,x\right)$
is uniformly bounded. Thus, we need to show the appropriate $\rho_{1}\left(\cdot\right)$
function. The following lemma shows it.
\begin{lemma}
The following functions satisfy (\ref{eq:lipschitz_rho1}) and (\ref{eq:rho1_to_0}).\end{lemma}
\begin{enumerate}
\item The function $\rho_{1}\left(y\right)=\left\Vert \ee_{2}-\ee_{1}\right\Vert $
and $\rho_{n2}\left(x\right)=\Gamma_{\eta}$ for $g_{n}\left(\ee,x\right)$.

\begin{enumerate}
\item The function $\rho_{1}\left(y\right)=\frac{1}{1-\lambda}B_{\phi}^{2}\left(\sum_{y\in\XS}B_{w}\left\Vert P\left(y|x,\theta_{1}\right)-P\left(y|x,\theta_{2}\right)\right\Vert +\left\Vert w_{1}-w_{2}\right\Vert \right)$
and $\rho_{n2}\left(x\right)=\Gamma_{w}$ for $g_{n}\left(w,x\right)$.
\item The function $\rho_{1}\left(y\right)=\sum_{y\in\XS}B_{w}\left\Vert P\left(y|x,\theta_{1}\right)-P\left(y|x,\theta_{2}\right)\right\Vert \cdot B_{\psi}$
and $\rho_{n2}\left(x\right)=1$ for $g_{n}\left(\theta,x\right)$.
\end{enumerate}
\end{enumerate}
\begin{proof}
$\quad$\end{proof}
\begin{enumerate}
\item Recalling (\ref{eq:g_n_eta}) we have for $g_{n}\left(\ee,x\right)$
\begin{eqnarray*}
\left\Vert g_{n}\left(\ee_{1},x\right)-g_{n}\left(\ee_{2},x\right)\right\Vert  & \le & \Gamma_{\eta}\left\Vert \ee_{2}-\ee_{1}\right\Vert ,\end{eqnarray*}
thus (\ref{eq:lipschitz_rho1}) and (\ref{eq:rho1_to_0}) are satisfied
for 1.
\item Recalling (\ref{eq:g_n_w}) we have for $g_{n}\left(w,x\right)$ \begin{eqnarray*}
\left\Vert g_{n}\left(w_{1},x\right)-g_{n}\left(w_{2},x\right)\right\Vert  & \le & \left\Vert \Gamma_{w}\sum_{k}^{\infty}\lambda^{k}\phi\left(x_{n-k}\right)\left(\left(\sum_{y\in\XS}P\left(y|x,\theta_{1}\right)\phi\left(y\right)'w_{1}-\phi\left(x_{n}\right)'w_{1}\right)\right.\right.\\
 &  & \left.\left.-\left(\sum_{y\in\XS}P\left(y|x,\theta_{2}\right)\phi\left(y\right)'w_{2}-\phi\left(x_{n}\right)'w_{2}\right)\right)\right\Vert \\
 & \le & \frac{\Gamma_{w}B_{\phi}^{2}}{1-\lambda}\left(\sum_{y\in\XS}\left\Vert P\left(y|x,\theta_{1}\right)w_{1}-P\left(y|x,\theta_{2}\right)w_{2}\right\Vert +\left\Vert w_{1}-w_{2}\right\Vert \right)\\
 & \le & \frac{\Gamma_{w}B_{\phi}^{2}}{1-\lambda}\left(\sum_{y\in\XS}B_{w}\left\Vert P\left(y|x,\theta_{1}\right)-P\left(y|x,\theta_{2}\right)\right\Vert +\left\Vert w_{1}-w_{2}\right\Vert \right)\end{eqnarray*}

\begin{enumerate}
\item Trivially, with respect to $w$ (\ref{eq:lipschitz_rho1}) and (\ref{eq:rho1_to_0})
are satisfied. Regarding $\theta$, (\ref{eq:lipschitz_rho1}) and
(\ref{eq:rho1_to_0}) are satisfied if we recall the definition of
$P\left(y|x,\theta\right)$ from (\ref{eq:trans_mat_policy}) and
the continuity of $\mu\left(u|x,\theta\right)$ from Assumption \ref{asum:mu_bound}.
\item Recalling (\ref{eq:g_n_theta}) we have for $g_{n}\left(\theta,x\right)$\begin{eqnarray*}
\left\Vert g_{n}\left(\theta_{1},x\right)-g_{n}\left(\theta_{2},x\right)\right\Vert  & = & \left\Vert \E\left[\left.\de\left(x,y,w_{1}\right)\psi\left(x,u,\theta_{1}\right)\right|\F_{n}\right]-\E\left[\left.\de\left(x,y,w_{2}\right)\psi\left(x,u,\theta_{2}\right)\right|\F_{n}\right]\right\Vert \\
 & \le & \sum_{y\in\XS}B_{w}\left\Vert P\left(y|x,\theta_{1}\right)-P\left(y|x,\theta_{2}\right)\right\Vert \cdot B_{\psi}.\end{eqnarray*}
Using similar arguments to 2, (\ref{eq:lipschitz_rho1}) and (\ref{eq:rho1_to_0})
are satisfied for $\theta$.
\end{enumerate}
\end{enumerate}

\section{Proof of Theorem \ref{thrm:main2_ODE_convergence}\label{app:main2_ODE_convergence}}

In this section we find conditions under which Algorithm \ref{algo:TD1ACFunction}
converges to a neighborhood of a local maximum. More precisely, we
show that $\liminf_{t\rightarrow\infty}\left\Vert \nabla\eta(\theta(t))\right\Vert _{2}\le\epsilon_{\textrm{app}}+\epsilon_{\textrm{dyn}}$,
where the approximation error, $\epsilon_{\textrm{app}}$, measures
the error inherent in the critic's representation, and $\epsilon_{\textrm{dyn}}$
is an error related to the single time scale algorithm. We note that
the approximation error depends on the basis functions chosen for
the critic, and in general can be reduced only by choosing a better
representation basis. The term $\epsilon_{\textrm{dyn}}$ is the dynamic
error, and this error can be reduced by choosing the critic's parameters
$\Gamma_{\eta}$ and $\Gamma_{w}$ appropriately.

We begin by establishing a variant of Lyapunov's theorem for asymptotic
stability%
\footnote{We say that the equilibrium point $x=0$ of the system $\dot{x}$=f$\left(x\right)$
is \textit{stable} if for each $\epsilon>0$ there exists a $\delta>0$
such that $\left\Vert x\left(0\right)\right\Vert <\delta\Rightarrow\left\Vert x\left(t\right)\right\Vert <\epsilon$
for all $t\ge0$. We say that the point $x=0$ is \textit{asymptotically
stable} if it is stable and there exists a $\delta>0$ such that $\left\Vert x\left(0\right)\right\Vert <\delta$
implies $\lim_{t\rightarrow\infty}x\left(t\right)=0$ (see \cite{Khalil02}
for more details). %
}, where instead of proving asymptotic convergence to a point, we prove
convergence to a compact invariant set. Based on this result, we continue
by establishing a bound on a time dependent ODE of the first order.
This result is used to bound the critic's error in estimating the
average reward per stage and the differential values. Finally, using
these results, we establish Theorem \ref{thrm:main2_ODE_convergence}.

We denote a closed ball of radius $y$ in some normed vector space,
$(\R^{L},\|\cdot\|_{2})$, by $\mathcal{B}_{y}$, and its surface
by $\partial\mathcal{B}_{y}$. Also, we denote by $A\backslash B$
a set, which contains all the members of set $A$ which are not members
of $B$. Finally, we define the complement of $\mathcal{B}_{y}$ by
$\mathcal{B}_{y}^{c}=\R^{L}\backslash\mathcal{B}_{y}$.

The following lemma is similar to Lyapunov's classic theorem for asymptotic
stability (\cite{Khalil02}, Theorem 4.1). The main difference is
that when the value of the Lyapunov function is unknown inside a ball,
convergence can be established to the ball, rather than to a single
point.
\begin{lemma}
\label{lemma:Lyapunov_variant} Consider a dynamical system, $\dot{x}=f\left(x\right)$
in a normed vector space, $(\R^{L},\|\cdot\|)$, and a closed ball
$\mathcal{B}_{r}\triangleq\left\{ x\left|x\in\R^{L},\|x\|\le r\right.\right\} $.
Suppose that there exists a continuously differentiable scalar function
$V\left(x\right)$ such that $V\left(x\right)>0$ and $\dot{V}\left(x\right)<0$
for all $x\in\mathcal{B}_{r}^{c}$, and $V\left(x\right)=0$ for $x\in\partial\mathcal{B}_{r}$.
Then, \[
\limsup_{t\rightarrow\infty}\|x\left(t\right)\|\le r.\]
 \end{lemma}
\begin{proof}
We prove two complementary cases. In the first case, we assume that
$x\left(t\right)$ never enters $\mathcal{B}_{r}$. On the set $\mathcal{B}_{r}^{c}$,
$V\left(x\right)$ is a strictly decreasing function in $t$, and
it is bounded below, thus it converges. We denote this bound by $C$,
and notice that $C\ge0$ since for $x\in\mathcal{B}_{r}^{c}$, $V\left(x\right)>0$.
We prove that $C=0$ by contradiction. Assume that $C>0$. Then, $x(t)$
converge to the invariant set $S_{C}\triangleq\{x|V\left(x\right)=C,x\in\mathcal{B}_{r}^{c}\}$.
For each $x\left(t\right)\in\mathcal{S}_{C}$ we have $\dot{V}\left(x\right)<0$.
Thus, $V\left(x\right)$ continues to decrease which contradicts the
boundedness from below. As a result, $V(x\left(t\right))\rightarrow0$.

In the second case, let us suppose that at some time, denoted by $t_{0}$,
$x(t_{0})\in\mathcal{B}_{r}$. We argue that the trajectory never
leaves $\mathcal{B}_{r}$. Let us assume that at some time $t_{2}$,
the trajectory $x(t)$ enters the set $\partial\mathcal{B}_{r+\epsilon}$.
Then on this set, we have $V(x(t_{2}))>0$. By the continuity of the
trajectory $x(t)$, the trajectory must go through the set $\partial\mathcal{B}_{r}$.
Denote the hitting time of this set by $t_{1}$. By definition we
have $V(x(t_{1}))=0$. Without loss of generality, we assume that
the trajectory in the times $t_{1}<t\le t_{2}$ is restricted to the
set $\mathcal{B}_{r+\epsilon}/\mathcal{B}_{r}$. Thus, since $\dot{V}(x(t))\le0$
for $x\in\mathcal{B}_{r+\epsilon}/\mathcal{B}_{r}$ we have \[
V(x(t_{2}))=V(x(t_{1}))+\int_{t_{1}}^{t_{2}}\dot{V}(x(t))dt<V(x(t_{1})),\]
 which contradicts the fact that $V(x(t_{2}))\ge V(x(t_{1}))$. Since
this argument holds for all $\epsilon>0$, the trajectory $x(t)$
never leaves $\mathcal{B}_{r}$.
\end{proof}
The following lemma will be applied later to the linear equations
\eqref{eq:ODE_w}, and more specifically, to the ODEs describing the
dynamics of $\tilde{\eta}$ and $w$. It bounds the difference between
an ODE's state variables and some time dependent functions.
\begin{lemma}
\label{lemma:boundedness_ODE} Consider the following ODE in a normed
space $(\R^{L},\|\cdot\|_{2})$\begin{equation}
\left\{ \begin{split}\frac{d}{dt}X\left(t\right) & =\M\left(t\right)\left(X\left(t\right)-F_{1}(t)\right)+F_{2}(t),\\
X(0) & =X_{0},\end{split}
\right.\label{eq:ODE_lemma}\end{equation}
where for sufficiently large $t$ .
\begin{enumerate}
\item $\M(t)\in\R^{L\times L}$ is a continuous matrix which satisfies $\max_{\left\Vert x\right\Vert =1}x'\M\left(t\right)x\le-\gamma<0$
for $t\in\mathbb{R}$,
\item $F_{1}\left(t\right)\in\R^{L}$ satisfies $\|dF_{1}(t)/dt\|_{2}\le B_{F1}$,
\item $F_{2}\left(t\right)\in\R^{L}$ satisfies $\|F_{2}(t)\|_{2}\le B_{F2}$.
\end{enumerate}
Then, the solution of the ODE satisfies $\limsup_{t\rightarrow0}\|X(t)-F_{1}\left(t\right)\|_{2}\le\left(B_{F1}+B_{F2}\right)/\gamma$.

\end{lemma}
\begin{proof}
We express \eqref{eq:ODE_lemma} as \begin{equation}
\frac{d}{dt}\left(X(t)-F_{1}(t)\right)=\M(t)\left(X(t)-F_{1}\left(t\right)\right)-\frac{d}{dt}F_{1}(t)+F_{2}(t),\label{eq:ODE_lemma_first_trick}\end{equation}
 and define \[
Z(t)\triangleq\left(X(t)-F_{1}(t)\right),\quad G(t)\triangleq-\frac{d}{dt}F_{1}(t)+F_{2}(t).\]
 Therefore, \eqref{eq:ODE_lemma_first_trick} can be written as \[
\dot{Z}(t)=\M(t)Z(t)+G(t),\]
 where $\|G(t)\|\le B_{G}\triangleq B_{F1}+B_{F2}$. In view of Lemma
\ref{lemma:Lyapunov_variant}, we consider the function \[
V\left(Z\right)=\frac{1}{2}\left(\left\Vert Z(t)\right\Vert _{2}^{2}-B_{G}^{2}/\gamma^{2}\right).\]
 Let $\mathcal{B}_{r}$ be a ball with a radius $r=B_{G}/\gamma$.
Thus we have $V\left(Z\right)>0$ for $Z\in\mathcal{B}_{r}^{c}$ and
$V(Z)=0$ for $X\in\partial\mathcal{B}_{r}$. In order to satisfy
the assumptions of Lemma \ref{lemma:Lyapunov_variant} the condition
that $\dot{V}(Z)<0$ needs to be verified. For $\left\Vert Z(t)\right\Vert _{2}>B_{G}/\gamma$
we have

\[
\begin{split}\dot{V}(Z) & =\left(\nabla_{X}V\right)'\dot{Z}(t)\\
 & =Z(t)'\M(t)Z(t)+Z(t)'G(t)\\
 & =\|Z(t)\|_{2}^{2}\frac{Z(t)'}{\|Z(t)\|_{2}}\M(t)\frac{Z(t)}{\|Z(t)\|_{2}}+Z(t)'G(t)\\
 & \le\|Z(t)\|_{2}^{2}\max_{\|Y(t)\|_{2}=1}Y(t)'\M(t)Y(t)+\left\Vert Z(t)\right\Vert _{2}\left\Vert G(t)\right\Vert _{2}\\
 & =\|Z(t)\|_{2}\left(-\gamma\left\Vert Z(t)\right\Vert _{2}+B_{G}\right)\\
 & <0.\end{split}
\]
As a result, the assumptions of Lemma \ref{lemma:Lyapunov_variant}
are valid and the Lemma is proved.
\end{proof}
The following lemma shows that the matrix $A\left(\theta\right)$,
defined in \eqref{eq:thrm_main_vars_w}, satisfies the conditions
of Lemma \ref{lemma:boundedness_ODE}. For the following lemmas, we
define the weighted norm $\left\Vert w\right\Vert _{\Pi\left(\theta\right)}^{2}\triangleq\left\Vert w'\Pi\left(\theta\right)w\right\Vert _{2}$.
\begin{lemma}
\label{lem:A_mat}The following inequalities hold:\end{lemma}
\begin{enumerate}
\item For any $w\in\R^{L}$and for all $\theta\in\R^{K}$, $\left\Vert P\left(\theta\right)w\right\Vert _{\Pi\left(\theta\right)}<\left\Vert w\right\Vert _{\Pi\left(\theta\right)}$..

\begin{enumerate}
\item The matrix $M\left(\theta\right)$ satisfies $\left\Vert M\left(\theta\right)w\right\Vert _{\Pi\left(\theta\right)}<\left\Vert w\right\Vert _{\Pi\left(\theta\right)}$
for all $\theta\in\R^{K}$ and $w\in\R^{L}$.
\item The matrix $\Pi\left(\theta\right)\left(M\left(\theta\right)-I\right)$
satisfies $x'\Pi\left(\theta\right)\left(M\left(\theta\right)-I\right)x<0$
for all $x\in\R^{L}$ and for all $\theta\in\R^{K}$.
\item There exists a positive scalar $\gamma$ such that $w'A\left(\theta\right)w<-\gamma$
for all $w'w=1$.
\end{enumerate}
\end{enumerate}
\begin{proof}
The following proof is similar in many aspects to the proof of Lemma
6.6 of \cite{BerTsi96}. \end{proof}
\begin{enumerate}
\item By using Jensen's inequality for the function $f\left(\alpha\right)=\alpha^{2}$
we have \begin{equation}
\left(\sum_{y\in\XS}P\left(y|x,\theta\right)w\left(y\right)\right)^{2}\le\sum_{y\in\XS}P\left(y|x,\theta\right)w\left(y\right)^{2},\quad\forall x\in\XS.\label{eq:Jensens}\end{equation}
If in Jensen's inequality we have a strictly convex fiction and non-degenerate
probability measures then the inequality is strict. The function $f\left(\alpha\right)$
is strictly convex, and by Assumption \ref{asum:aperiodicy_reccurent}
the matrix $P\left(\theta\right)$ is aperiodic, which implies that
the matrix $P\left(\theta\right)$ is not a permutation matrix. As
a result, there exists $x_{0}\in\XS$ such that the probability measure
$P\left(y|x_{0},\theta\right)$ is not degenerate, thus, the inequality
in \eqref{eq:Jensens} is strict, i.e., \begin{equation}
\left(\sum_{y\in\XS}P\left(y|x_{0},\theta\right)w\left(y\right)\right)^{2}<\sum_{y\in\XS}P\left(y|x_{0},\theta\right)w\left(y\right)^{2}.\label{eq:Jensen_strict}\end{equation}
Then, we have \begin{eqnarray*}
\left\Vert P\left(\theta\right)w\right\Vert _{\Pi\left(\theta\right)} & = & w'P\left(\theta\right)'\Pi\left(\theta\right)P\left(\theta\right)w\\
 & = & \sum_{x\in\XS}\pi\left(x|\theta\right)\left(\sum_{y\in\XS}P\left(y|x,\theta\right)w\left(y\right)\right)^{2}\\
 & < & \sum_{x\in\XS}\pi\left(x|\theta\right)\sum_{y\in\XS}P\left(y|x,\theta\right)w\left(y\right)^{2}\\
 & = & \sum_{y\in\XS}w\left(y\right)^{2}\sum_{x\in\XS}\pi\left(x|\theta\right)P\left(y|x,\theta\right)\\
 & = & \sum_{y\in\XS}w\left(y\right)^{2}\pi\left(y|\theta\right)\\
 & = & \left\Vert w\right\Vert _{\Pi\left(\theta\right)},\end{eqnarray*}
where in the inequality we have used \eqref{eq:Jensen_strict}.

\begin{enumerate}
\item Using the triangle inequality and 1 we have\begin{eqnarray*}
\left\Vert M\left(\theta\right)w\right\Vert _{\Pi\left(\theta\right)} & = & \left\Vert \left(1-\lambda\right)\sum_{m=0}^{\infty}\lambda^{m}P\left(\theta\right)^{m+1}w\right\Vert _{\Pi\left(\theta\right)}\\
 & \le & \left(1-\lambda\right)\sum_{m=0}^{\infty}\lambda^{m}\left\Vert P\left(\theta\right)^{m+1}w\right\Vert _{\Pi\left(\theta\right)}\\
 & < & \left(1-\lambda\right)\sum_{m=0}^{\infty}\lambda^{m}\left\Vert w\right\Vert _{\Pi\left(\theta\right)}\\
 & = & \left\Vert w\right\Vert _{\Pi\left(\theta\right)}.\end{eqnarray*}

\item By definition\begin{eqnarray*}
x'\Pi\left(\theta\right)M\left(\theta\right)x & = & x'\Pi\left(\theta\right)^{1/2}\Pi\left(\theta\right)^{1/2}M\left(\theta\right)x\\
 & \le & \left\Vert \Pi\left(\theta\right)^{1/2}x\right\Vert \cdot\left\Vert \Pi\left(\theta\right)^{1/2}M\left(\theta\right)x\right\Vert \\
 & = & \left\Vert x\right\Vert _{\Pi\left(\theta\right)}\left\Vert M\left(\theta\right)x\right\Vert _{\Pi\left(\theta\right)}\\
 & < & \left\Vert x\right\Vert _{\Pi\left(\theta\right)}\left\Vert x\right\Vert _{\Pi\left(\theta\right)}\cdot\\
 & = & x'\Pi\left(\theta\right)x,\end{eqnarray*}
where in the first inequality we have used the Cauchy-Schwartz inequality,
and in the second inequality we have used 1. Thus, $x'\Pi\left(\theta\right)\left(M\left(\theta\right)-I\right)x<0$
for all $x\in\mathbb{R},$ which implies that $\Pi\left(\theta\right)\left(M\left(\theta\right)-I\right)$
is a negative definite (ND) matrix%
\footnote{Usually, a ND matrix is defined for Hermitian matrices, i.e., if $B$
is an Hermitian matrix and it satisfies $x'Bx<0$ for all $x\in\mathbb{C}^{K}$
then $B$ is a NSD matrix . We use here a different definition which
states that a square matrix $B$ is a ND matrix if it is real and
it satisfies $x'Bx<0$ for all $x\in\mathbb{R}^{k}$ (see \cite{HornJohnson}
p. 399).%
}.
\item From 3, we know that for all $\theta\in\R^{K}$ and all $w\in\R^{\left|\XS\right|}$
satisfying $w'w=1$, we have $w'\Pi\left(\theta\right)\left(M\left(\theta\right)-I\right)w<0$,
and by Assumption \eqref{asum:aperiodicy_reccurent}, this is true
also for the closure of $\left\{ \Pi\left(\theta\right)\left(M\left(\theta\right)-I\right)|\theta\in\R^{K}\right\} $.
Thus, there exists a positive scalar, $\gamma'$, satisfying \[
w'\Pi\left(\theta\right)\left(M\left(\theta\right)-I\right)w\le-\gamma'<0.\]
By Assumption \ref{asum:Phi_linear_indepen_bound} the rank of the
matrix $\Phi$ is full, thus there exists a scalar $\gamma$ such
that for all $w\in\R^{L}$, where $w'w=1$, we have $w'A\left(\theta\right)w\le-\gamma<0$.
\end{enumerate}
\end{enumerate}
The following Lemma establishes the boundedness of $\dot{\theta}$.
\begin{lemma}
\label{lem:bounded_theta_D1}There exists a constant $B_{\theta1}\triangleq B_{\eta1}+B_{\psi}\left(B_{D}+B_{r}+B_{\tilde{\eta}}+2B_{\phi}B_{w}\right)$
such that $\|\dot{\theta}\|_{2}\le B_{\theta1}$. \end{lemma}
\begin{proof}
Recalling \eqref{eq:ODE_w} \[
\begin{split}\left\Vert \dot{\theta}\right\Vert _{2} & =\left\Vert \nabla_{\theta}\eta(\theta)+\sum_{x,y\in\XS\times\XS,u\in\US}D^{(x,u,y)}(\theta)\left(d(x,y,\theta)-\tilde{d}(x,y,w)\right)\right\Vert _{2}\\
 & \le B_{\eta1}+\sum_{x,y\in\XS\times\XS,u\in\US}\left\Vert D^{(x,u,y)}(\theta)\right\Vert _{2}\left\Vert d(x,y,\theta)-\tilde{d}(x,y,w)\right\Vert _{2}\\
 & \le B_{\eta1}+B_{\psi}\left(B_{D}+B_{r}+B_{\tilde{\eta}}+2B_{\phi}B_{w}\right)\\
 & \triangleq B_{\theta1}.\end{split}
\]

\end{proof}
Based on Lemma \eqref{lem:bounded_theta_D1}, the following Lemma
shows the boundedness of $\left(\eta(\theta(t))-\EtaE\right)$.
\begin{lemma}
\label{lemma:bound_eta-eta} We have \[
\limsup_{t\rightarrow\infty}\left|\eta(\theta(t))-\EtaE\right|\le\frac{B_{\Delta\eta}}{\Gamma_{\eta}},\]
where $B_{\Delta\eta}\triangleq B_{\eta_{1}}B_{\theta1}$.\end{lemma}
\begin{proof}
Using the Cauchy-Schwartz inequality we have \begin{equation}
\begin{split}|\dot{\eta}(\theta)| & =|\nabla\eta(\theta)'\dot{\theta}|\\
 & \le\left\Vert \nabla\eta(\theta)\right\Vert _{2}\|\dot{\theta}\|_{2}\\
 & \le B_{\eta_{1}}B_{\theta1}.\end{split}
\label{eq:eta_dot_bounded}\end{equation}
 Recalling the equation for $\EtaE$ in \eqref{eq:ODE_w} we have
\[
\dot{\EtaE}=\Gamma_{\eta}\left(\eta(\theta)-\EtaE\right).\]
We conclude by applying Lemma \ref{lemma:boundedness_ODE} and using
(\ref{eq:eta_dot_bounded}) that \begin{equation}
\limsup_{t\rightarrow\infty}\left|\eta(\theta(t))-\EtaE\right|\le\frac{B_{\eta_{1}}B_{\theta1}}{\Gamma_{\eta}}=\frac{B_{\Delta\eta}}{\Gamma_{\eta}}.\label{eq:eta_minus_eta_bounded}\end{equation}

\end{proof}
In \eqref{eq:eta_minus_eta_bounded} we see that the bound on $|\eta(\theta)-\EtaE|$
is controlled by $\Gamma_{\eta}$, where larger values of $\Gamma_{\eta}$
ensure smaller values of $|\eta(\theta)-\EtaE|$. Next, we bound $\|w^{*}(\theta)-w\|_{2}$.
We recall the second equation of \eqref{eq:ODE_w} \begin{eqnarray*}
\dot{w} & = & \Psi_{w}\left[\Gamma_{w}\left(A\left(\theta\right)w+b\left(\theta\right)+G(\theta)(\eta(\theta)-\ee)\right)\right],\\
A\left(\theta\right) & = & \Phi'\Pi\left(\theta\right)\left(M-I\right)\Phi,\\
M\left(\theta\right) & = & \left(1-\lambda\right)\sum_{m=0}^{\infty}\lambda^{m}P\left(\theta\right)^{m+1},\\
b\left(\theta\right) & = & \Phi'\Pi\left(\theta\right)\sum_{m=0}^{\infty}\lambda^{m}P\left(\theta\right)^{m}\left(r-\eta\left(\theta\right)\right),\\
G(\theta) & = & \Phi'\Pi\left(\theta\right)\sum_{m=0}^{\infty}\lambda^{m}P\left(\theta\right)^{m}.\end{eqnarray*}
We can write the equation for $\dot{w}$ as\[
\dot{w}=\Psi_{w}\left[\Gamma_{w}\left(A\left(\theta\right)\left(w-w^{*}\left(\theta\right)\right)+G(\theta)(\eta(\theta)-\ee)\right)\right],\]
where $w^{*}=-A\left(\theta\right)^{-1}b\left(\theta\right)$. In
order to use Lemma \ref{lemma:boundedness_ODE}, we need to demonstrate
the boundedness of $\left\Vert \frac{d}{dt}w^{*}\right\Vert $. The
following lemma does so. \\

\begin{lemma}
\label{lem:w_dot_bounded}$ $\end{lemma}
\begin{enumerate}
\item There exists a positive constant, $B_{b}\triangleq\frac{1}{1-\lambda}\left|\XS\right|^{3}LB_{\Phi}B_{r}$,
such that $\left\Vert b\left(\theta\right)\right\Vert _{2}\le B_{b}$.

\begin{enumerate}
\item There exists a positive constant, $B_{G}\triangleq\frac{1}{1-\lambda}\left|\XS\right|^{3}LB_{\Phi}$,
such that $\left\Vert G\left(\theta\right)\right\Vert _{2}\le B_{G}$.
\item There exist positive constants, $\tilde{B}=B_{\pi1}\left(B_{r}+B_{\eta}\right)B_{\theta1}+B_{P1}\left(B_{r}+B_{\eta}\right)B_{\theta1}+B_{\eta1}B_{\theta1}$
and $B_{b1}\triangleq\frac{1}{1-\lambda}\left|\XS\right|^{3}B_{\Phi}B_{r}\tilde{B}$,
such that we have $\left\Vert \dot{b}\left(\theta\right)\right\Vert _{2}\le B_{b1}.$
\item There exist constants $b_{A}$ and $B_{A}$ such that\[
0<b_{A}\le\left\Vert A\left(\theta\right)\right\Vert _{2}\le B_{A}.\]

\item There exist a constants $B_{A1}$ such that\[
\left\Vert A\left(\theta\right)\right\Vert _{2}\le B_{A1}.\]

\item We have \[
\left\Vert \frac{d}{dt}\left(A\left(\theta\right)^{-1}\right)\right\Vert _{2}\le b_{A}^{2}B_{A1}.\]

\item There exists a positive constant, $B_{w1}$, such that\[
\left\Vert \frac{d}{dt}w^{*}\right\Vert _{2}\le B_{w1}.\]

\end{enumerate}
\end{enumerate}
\begin{proof}
$ $\end{proof}
\begin{enumerate}
\item We show that the entries of the vector $b\left(\theta\right)$ are
uniformly bounded in $\theta$, therefore, its norm is uniformly bounded
in $\theta$. Let us look at the $i$-th entry of the vector $b\left(\theta\right)$
(we denote by $\left[\cdot\right]_{j}$ the $j$-th row of a matrix
or a vector) \begin{eqnarray*}
\left|\left[b\left(\theta\right)\right]_{i}\right| & = & \left|\left[\Phi'\Pi\left(\theta\right)\sum_{m=0}^{\infty}\lambda^{m}P\left(\theta\right)^{m}\left(r-\eta\left(\theta\right)\right)\right]_{i}\right|\\
 & \le & \sum_{m=0}^{\infty}\lambda^{m}\left|\left[\Phi'\Pi\left(\theta\right)P\left(\theta\right)^{m}\left(r-\eta\left(\theta\right)\right)\right]_{i}\right|\\
 & = & \sum_{m=0}^{\infty}\lambda^{m}\left|\sum_{l=1}^{\left|\XS\right|}\sum_{j=1}^{\left|\XS\right|}\sum_{k=1}^{\left|\XS\right|}\left[\Phi'\right]_{ik}\Pi_{kj}\left(\theta\right)\left[P\left(\theta\right)^{m}\right]_{jl}\left(r_{l}-\eta\left(\theta\right)\right)\right|\\
 & \le & \frac{1}{1-\lambda}\left|\XS\right|^{3}B_{\Phi}B_{r},\end{eqnarray*}
thus $\left\Vert b\left(\theta\right)\right\Vert _{2}\le\frac{1}{1-\lambda}\left|\XS\right|^{3}LB_{\Phi}B_{r}$
is uniformly bounded in $\theta$.
\item The proof is accomplished by similar argument to section 1.
\item Similarly to section 1, we show that the entries of the vector $\dot{b}\left(\theta\right)$
are uniformly bounded in $\theta$, therefore, its norm is uniformly
bounded in $\theta$. First, we show that the following function of
$\theta\left(t\right)$ is bounded.\begin{eqnarray*}
\left|\frac{d}{dt}\left(\Pi_{kj}\left(\theta\right)\left[P\left(\theta\right)^{m}\right]_{jl}\left(r_{l}-\eta\left(\theta\right)\right)\right)\right| & = & \left|\nabla_{\theta}\left(\Pi_{kj}\left(\theta\right)\left[P\left(\theta\right)^{m}\right]_{jl}\left(r_{l}-\eta\left(\theta\right)\right)\right)\dot{\theta}\right|\\
 & \le & \left|\left(\nabla_{\theta}\Pi_{kj}\left(\theta\right)\right)\left[P\left(\theta\right)^{m}\right]_{jl}\left(r_{l}-\eta\left(\theta\right)\right)\dot{\theta}\right|\\
 &  & +\left|\Pi_{kj}\left(\theta\right)\left[\nabla_{\theta}P\left(\theta\right)^{m}\right]_{jl}\left(r_{l}-\eta\left(\theta\right)\right)\dot{\theta}\right|\\
 &  & +\left|\Pi_{kj}\left(\theta\right)\left[P\left(\theta\right)^{m}\right]_{jl}\nabla_{\theta}\left(r_{l}-\eta\left(\theta\right)\right)\dot{\theta}\right|\\
 & \le & B_{\pi1}\left(B_{r}+B_{\eta}\right)\cdot B_{\theta1}+B_{P1}\left(B_{r}+B_{\eta}\right)B_{\theta1}+B_{\eta1}B_{\theta1}\\
 & = & \tilde{B},\end{eqnarray*}
where we used the triangle and Cauchy-Schwartz inequalities in the
first and second inequalities respectively, and Lemmas \ref{lemma:pi_eta_P_bounded}
and \ref{lem:bounded_theta_D1} in the second inequality. Thus, \begin{eqnarray*}
\left|\left[\dot{b}\left(\theta\right)\right]_{i}\right| & = & \left|\left[\Phi'\Pi\left(\theta\right)\sum_{m=0}^{\infty}\lambda^{m}P\left(\theta\right)^{m}\left(r-\eta\left(\theta\right)\right)\right]_{i}\right|\\
 & \le & \sum_{m=0}^{\infty}\lambda^{m}\left|\left[\Phi'\Pi\left(\theta\right)P\left(\theta\right)^{m}\left(r-\eta\left(\theta\right)\right)\right]_{i}\right|\\
 & = & \sum_{m=0}^{\infty}\lambda^{m}\left|\sum_{l=1}^{\left|\XS\right|}\sum_{j=1}^{\left|\XS\right|}\sum_{k=1}^{\left|\XS\right|}\left[\Phi'\right]_{ik}\frac{d}{dt}\left(\Pi_{kj}\left(\theta\right)\left[P\left(\theta\right)^{m}\right]_{jl}\left(r_{l}-\eta\left(\theta\right)\right)\right)\right|\\
 & \le & \frac{1}{1-\lambda}\left|\XS\right|^{3}B_{\Phi}B_{r}\tilde{B}\\
 & = & B_{b1}.\end{eqnarray*}

\item Since $A(\theta)$ satisfies $y'A(\theta)y<0$ for all nonzero $y$,
it follows that all its eigenvalues are nonzero. Therefore, the eigenvalues
of $A\left(\theta\right)'A\left(\theta\right)$ are all positive and
real since $A\left(\theta\right)'A\left(\theta\right)$ is a symmetric
matrix. Since by Assumption \ref{asum:aperiodicy_reccurent} this
holds for all $\theta\in\R^{K}$, there is a global minimum, $b_{A}$,
and a global maximum, $B_{A}$, such that\[
B_{A}^{2}\ge\lambda_{\textrm{max}}\left(A\left(\theta\right)'A\left(\theta\right)\right)\ge\lambda_{\textrm{min}}\left(A\left(\theta\right)'A\left(\theta\right)\right)\ge b_{A}^{2},\quad\forall\theta\in\R^{K},\]
where we denote by $\lambda_{\textrm{min}}\left(\cdot\right)$ and
$\lambda_{\textrm{max}}\left(\cdot\right)$ the minimal and maximal
eigenvalues of the matrix respectively. Using \cite{HornJohnson}
section 5.6.6, we have $\lambda_{max}\left(A\left(\theta\right)'A\left(\theta\right)\right)=\left\Vert A\left(\theta\right)\right\Vert _{2}$,
thus, we get an upper bound on the matrix norm. Let us look at the
norm of $\left\Vert A\left(\theta\right)^{-1}\right\Vert _{2}$,\begin{eqnarray*}
\left\Vert A\left(\theta\right)^{-1}\right\Vert _{2}^{2} & = & \lambda_{\mathrm{max}}\left(\left(A\left(\theta\right)^{-1}\right)'A\left(\theta\right)^{-1}\right)\\
 & = & \lambda_{\mathrm{max}}\left(\left(A\left(\theta\right)'\right)^{-1}A\left(\theta\right)^{-1}\right)\\
 & = & \lambda_{\mathrm{max}}\left(\left(A\left(\theta\right)A\left(\theta\right)'\right)^{-1}\right)\\
 & = & 1/\lambda_{\mathrm{min}}\left(A\left(\theta\right)A\left(\theta\right)'\right)\\
 & = & 1/\lambda_{\mathrm{min}}\left(\left(A\left(\theta\right)'A\left(\theta\right)\right)'\right)\\
 & = & 1/\lambda_{\mathrm{min}}\left(A\left(\theta\right)'A\left(\theta\right)\right),\end{eqnarray*}
thus, we the lower bound on $\left\Vert A\left(\theta\right)^{-1}\right\Vert _{2}$
is $\sqrt{1/\lambda_{\mathrm{min}}\left(A\left(\theta\right)'A\left(\theta\right)\right)}$,
i.e., $b_{A}$.
\item Let us look at the $ij$ entry of the matrix $\frac{d}{dt}A\left(\theta\right)$,
where using similar arguments to section 2 we get \begin{eqnarray*}
\left[\left|\frac{d}{dt}A\left(\theta\right)\right|\right]_{ij} & = & \left[\left|\frac{d}{dt}\left(\Phi'\Pi\left(\theta\right)\left(\left(1-\lambda\right)\sum_{m=0}^{\infty}\lambda^{m}P\left(\theta\right)^{m+1}-I\right)\Phi\right)\right|\right]_{ij}\\
 & \le & \left[\left|\Phi'\frac{d}{dt}\left(\Pi\left(\theta\right)\right)\left(\left(1-\lambda\right)\sum_{m=0}^{\infty}\lambda^{m}P\left(\theta\right)^{m+1}-I\right)\Phi\right|\right]_{ij}\\
 &  & +\left[\left|\Phi'\Pi\left(\theta\right)\frac{d}{dt}\left(\left(1-\lambda\right)\sum_{m=0}^{\infty}\lambda^{m}P\left(\theta\right)^{m+1}-I\right)\Phi\right|\right]_{ij}\\
 & \le & B_{\Phi}B_{\pi1}\frac{1}{1-\lambda}B_{\Phi}+B_{\Phi}\frac{1}{\left(1-\lambda\right)^{2}}B_{P1}B_{\Phi}.\end{eqnarray*}
Since the matrix entries are uniformly bounded in $\theta$, so is
the matrix $\frac{d}{dt}A\left(\theta\right)'\frac{d}{dt}A\left(\theta\right)$,
and so is the largest eigenvalue of $\frac{d}{dt}A\left(\theta\right)'\frac{d}{dt}A\left(\theta\right)$
which implies the uniform boundedness of $\left\Vert \frac{d}{dt}A\left(\theta\right)\right\Vert _{2}$.
\item For a general invertible square matrix, $X\left(t\right)$, we have\begin{eqnarray*}
0 & = & \frac{d}{dt}I=\frac{d}{dt}\left(X\left(t\right)^{-1}X\left(t\right)\right)=\frac{d}{dt}\left(X\left(t\right)^{-1}\right)X\left(t\right)+X\left(t\right)^{-1}\frac{d}{dt}\left(X\left(t\right)\right).\end{eqnarray*}
Rearranging it we get\[
\frac{d}{dt}\left(X\left(t\right)^{-1}\right)=-X\left(t\right)^{-1}\frac{d}{dt}\left(X\left(t\right)\right)X\left(t\right)^{-1}.\]
Using this identity yields\begin{eqnarray*}
\left\Vert \frac{d}{dt}\left(A\left(\theta\right)^{-1}\right)\right\Vert _{2} & = & \left\Vert -A\left(\theta\right)^{-1}\frac{d}{dt}\left(A\left(\theta\right)\right)A\left(\theta\right)^{-1}\right\Vert _{2}\\
 & \le & \left\Vert A\left(\theta\right)^{-1}\right\Vert _{2}\cdot\left\Vert \frac{d}{dt}\left(A\left(\theta\right)\right)\right\Vert _{2}\cdot\left\Vert -A\left(\theta\right)^{-1}\right\Vert _{2}\\
 & = & b_{A}^{2}B_{A1}.\end{eqnarray*}

\item Examining the norm of $\frac{d}{dt}w^{*}$ yields\begin{eqnarray*}
\left\Vert \frac{d}{dt}w^{*}\right\Vert _{2} & = & \left\Vert \frac{d}{dt}\left(A\left(\theta\right)^{-1}b\left(\theta\right)\right)\right\Vert _{2}\\
 & = & \left\Vert \frac{d}{dt}A\left(\theta\right)^{-1}b\left(\theta\right)+A\left(\theta\right)^{-1}\frac{d}{dt}b\left(\theta\right)\right\Vert _{2}\\
 & \le & b_{A}^{2}B_{A1}\frac{1}{1-\lambda}\left|\XS\right|^{3}B_{\Phi}B_{r}+b_{A}\tilde{B}\\
 & = & B_{w1}.\end{eqnarray*}

\end{enumerate}
We wish to use Lemma \ref{lemma:boundedness_ODE} for \eqref{eq:ODE_w},
thus, we show that the assumptions of Lemma \ref{lemma:boundedness_ODE}
are valid.
\begin{lemma}
\label{lemma:bound_w-w}$ $\end{lemma}
\begin{enumerate}
\item We have \begin{equation}
\limsup_{t\rightarrow\infty}\|w^{*}(\theta(t))-w(t)\|_{2}\le\frac{1}{\Gamma_{w}}B_{\Delta w},\label{eq:w_star-w_bound}\end{equation}
where \[
B_{\Delta w}\triangleq\frac{B_{w1}+B_{G}\frac{B_{\Delta\eta}}{\Gamma_{\eta}}}{\gamma}.\]

\begin{enumerate}
\item We have \[
\limsup_{t\rightarrow\infty}\|h(\theta(t))-\tilde{h}(w(t))\|_{\infty}\le\frac{B_{\Delta h1}}{\Gamma_{w}}+\frac{\epsilon_{\textrm{app}}}{\sqrt{b_{\pi}}},\]
where \[
B_{\Delta h}\triangleq|\XS|L\left(B_{\Delta w}\right)^{2}.\]

\end{enumerate}
\end{enumerate}
\begin{proof}
$ $\end{proof}
\begin{enumerate}
\item Without loss of generality, we can eliminate the projection operator
since we can choose $B_{w}$ to be large enough such that $w^{*}(\theta)$
will be inside the bounded space. We take $\M(t)=A\left(\theta\right)$,
$F_{1}\left(t\right)=w^{*}(\theta(t))$, and $F_{2}\left(t\right)=G(\theta)(\eta(\theta)-\ee)$
. By previous lemmas we can see that the Assumption \ref{lemma:boundedness_ODE}
holds. By Lemma \ref{lem:w_dot_bounded} (6), $\left\Vert \dot{w}^{*}(\theta)\right\Vert _{2}$
is bounded by $B_{w_{1}}$, by Lemma \ref{lemma:bound_eta-eta} we
have a bound on $|(\eta(\theta)-\EtaE)|$, and by Lemma \ref{lem:A_mat}
we have a bound on $w'A\left(\theta\right)w$. Using these bounds
and applying Lemma \ref{lemma:boundedness_ODE} provides the desired
result.

\begin{enumerate}
\item Suppressing the time dependence for simplicity and expressing $\|h(\theta)-\he(w)\|_{\infty}$
using $\epsilon_{\textrm{app}}$ and the previous result yields \begin{equation}
\begin{split}\|h(\theta)-\he(w)\|_{\infty} & \le\|h(\theta)-\he(w)\|_{2}\\
 & =\|h(\theta)-\he(w^{*})+\he(w^{*})-\he(w)\|_{2}\\
 & \le\|h(\theta)-\he(w^{*})\|_{2}+\|\he(w^{*})-\he(w)\|_{2}\end{split}
\label{eq:h_bound}\end{equation}
 For the first term on the r.h.s. of the final equation in \eqref{eq:h_bound}
we have \begin{equation}
\begin{split}\|h(\theta)-\he(w^{*})\|_{2} & =\left\Vert \left(\Pi(\theta)^{-\frac{1}{2}}\right)\left(\Pi(\theta)^{\frac{1}{2}}\right)\left(h(\theta)-\he(w^{*})\right)\right\Vert _{2}\\
 & \le\left\Vert \Pi(\theta)^{-\frac{1}{2}}\right\Vert _{2}\left\Vert h(\theta)-\he(w^{*})\right\Vert _{\Pi(\theta)}\\
 & \le\frac{\epsilon_{\textrm{app}}}{(b_{\pi})^{\frac{1}{2}}}\end{split}
\label{eq:approx_h}\end{equation}
where we use the sub-additivity of the matrix norms in the first inequality,
and Lemma \ref{lemma:pi_eta_P_bounded} and the \eqref{eq:FA_err_bound}
in the last inequality. For the second term on the r.h.s.\! of the
final equation in \eqref{eq:h_bound} we have \begin{equation}
\begin{split}\|\he(w^{*})-\he(w)\|_{2}^{2} & =\|\Phi(w^{*}(\theta)-w)\|_{2}^{2}\\
 & =\sum_{k=1}^{|\XS|}\left(\sum_{l=1}^{L}\phi_{l}(k)\left(w_{l}^{*}(\theta)-w_{l}\right)\right)^{2}\\
 & \le\sum_{k=1}^{|\XS|}\left(\left(\sum_{l=1}^{L}\phi_{l}^{2}(k)\right)^{\frac{1}{2}}\left(\sum_{l=1}^{L}\left(w_{l}^{*}(\theta)-w_{l}\right)^{2}\right)^{\frac{1}{2}}\right)^{2}\\
 & \le\sum_{k=1}^{|\XS|}\left(\sum_{l=1}^{L}\phi_{l}^{2}(k)\right)\left(\sum_{l=1}^{L}\left(w_{l}^{*}(\theta)-w_{l}\right)^{2}\right)\\
 & \le|\XS|L\|w^{*}(\theta)-w\|_{2}^{2}\\
 & =|\XS|L\left(B_{\Delta w}\right)^{2}.\end{split}
\label{eq:h(w_star)_bound}\end{equation}
 Combining \eqref{eq:w_star-w_bound}-\eqref{eq:h(w_star)_bound}
yields the desired result.
\end{enumerate}
\end{enumerate}
Using Lemma \ref{lemma:bound_w-w} we can provide a bound on second
term of \eqref{eq:ODE_w}.
\begin{lemma}
We have \[
\limsup_{t\rightarrow\infty}\left\Vert \sum_{x,y\in\XS\times\XS,u\in\US}D^{(x,u,y)}(\theta)\left(d(x,y,\theta)-\tilde{d}(x,y,w)\right)\right\Vert _{2}\le\frac{B_{\Delta td1}}{\Gamma_{w}}+\frac{B_{\Delta td2}}{\Gamma_{\eta}}+B_{\Delta td3}\epsilon_{\textrm{app}}\]
 where \[
B_{\Delta td1}=\frac{1}{\Gamma_{w}}\cdot2B_{\Psi}B_{\Delta h1},\quad B_{\Delta td2}=\frac{1}{\Gamma_{\eta}}\cdot B_{\Delta\eta}B_{\Psi},\quad B_{\Delta td3}=\frac{2B_{\Psi}}{\sqrt{b_{\pi}}}.\]
 \end{lemma}
\begin{proof}
Simplifying the notation by suppressing the time dependence, we bound
the TD signal in the limit, i.e., \[
\begin{split}\limsup_{t\rightarrow\infty}|d(x,y,\theta)-\tilde{d}(x,y,w)| & =\limsup_{t\rightarrow\infty}\left|\left(r(x)-\eta(\theta)+h(y,\theta)-h(x,\theta)\right)-\left(r(x)-\EtaE+\he(y,w)-\he(x,w)\right)\right|\\
 & \le\limsup_{t\rightarrow\infty}\left|\eta(\theta)-\EtaE\right|+\limsup_{t\rightarrow\infty}2\left\Vert h(\theta)-\he(w)\right\Vert _{\infty}\\
 & =\frac{B_{\Delta\eta}}{\Gamma_{\eta}}+2\left(\frac{B_{\Delta h1}}{\Gamma_{w}}+\frac{\epsilon_{\textrm{app}}}{\sqrt{b_{\pi}}}\right).\end{split}
\]
 With some more algebra we have \[
\begin{split} & \limsup_{t\rightarrow\infty}\left\Vert \sum_{x,y\in\XS\times\XS,u\in\US}D^{(x,u,y)}(\theta)\left(d(x,y,\theta)-\tilde{d}(x,y,w)\right)\right\Vert \\
 & \le\limsup_{t\rightarrow\infty}\sum_{x,y\in\XS\times\XS,u\in\US}\pi\left(x\right)P\left(u|x,\theta_{n}\right)P\left(y|x,u\right)\left\Vert \psi\left(x,u,\theta_{n}\right)\right\Vert \cdot\left|d(x,y,\theta)-\tilde{d}(x,y,w)\right|\\
 & \le B_{\Psi}\left(\frac{B_{\Delta\eta}}{\Gamma_{\eta}}+2\left(\frac{B_{\Delta h1}}{\Gamma_{w}}+\frac{\epsilon_{\textrm{app}}}{\sqrt{b_{\pi}}}\right)\right)\\
 & =\frac{B_{\Delta td1}}{\Gamma_{w}}+\frac{B_{\Delta td2}}{\Gamma_{\eta}}+B_{\Delta td3}\epsilon_{\textrm{app}}.\end{split}
\]

\end{proof}
We see that the term in this bound is adjustable by choosing appropriate
$\Gamma_{\eta}$ and $\Gamma_{w}$. The concluding lemma proves the
conclusion of Theorem \ref{thrm:main2_ODE_convergence}.

\subsection*{Proof of Theorem \ref{thrm:main2_ODE_convergence}}

\global\long\def\bd{B_{\nabla\eta}}
We define \[
B_{\nabla\eta}\triangleq\frac{B_{\Delta td1}}{\Gamma_{w}}+\frac{B_{\Delta td2}}{\Gamma_{\eta}}+B_{\Delta td3}\epsilon_{\textrm{app}}.\]
For an arbitrary $\delta>0$, define the set \begin{equation}
\mathcal{B}_{\delta}\triangleq\{\theta:~\left\Vert \nabla\eta(\theta)\right\Vert \le\bd+\delta\}.\end{equation}
 We claim that the trajectory $\eta(\theta)$ visits $\mathcal{B}_{\delta}$
infinitely often. Assume the contrary that \begin{equation}
\liminf_{t\rightarrow\infty}\left\Vert \nabla\eta(\theta)\right\Vert _{2}>\bd+\delta.\label{eq:contradiction_assum}\end{equation}
 Thus, on the set $\mathcal{B}_{\delta}^{c}$ for $t$ large enough
we have \begin{equation}
\begin{split}\dot{\eta}(\theta) & =\nabla\eta(\theta)\cdot\dot{\theta}\\
 & =\nabla\eta(\theta)\cdot\left(\nabla\eta(\theta)+\sum_{x,y\in\XS\times\XS}D^{(x,y)}(\theta)\left(d(x,y)-\tilde{d}(x,y)\right)\right)\\
 & =\left\Vert \nabla\eta(\theta)\right\Vert _{2}^{2}+\nabla\eta(\theta)\cdot\left(\sum_{x,y\in\XS\times\XS}D^{(x,y)}(\theta)\left(d(x,y)-\tilde{d}(x,y)\right)\right)\\
 & \ge\left\Vert \nabla\eta(\theta)\right\Vert _{2}^{2}-\left\Vert \nabla\eta(\theta)\right\Vert _{2}\left\Vert \sum_{x,y\in\XS\times\XS}D^{(x,y)}(\theta)\left(d(x,y)-\tilde{d}(x,y)\right)\right\Vert _{2}\\
 & =\left\Vert \nabla\eta(\theta)\right\Vert _{2}\left(\left\Vert \nabla\eta(\theta)\right\Vert _{2}-\bd\right)\\
 & \ge\left\Vert \nabla\eta(\theta)\right\Vert _{2}\left(\bd+\delta-\bd\right)\\
 & >(\bd+\delta)\delta.\end{split}
\label{eq:dot_eta_theta_1}\end{equation}
 By \eqref{eq:contradiction_assum}, there exists a time $t_{0}$
which for all $t>t_{0}$ we have $\eta(\theta)\in\mathcal{B}_{\delta}^{c}$.
Therefore, \begin{equation}
\eta(\infty)=\eta(t_{0})+\int_{t_{0}}^{\infty}\dot{\eta}(\theta)dt>\eta(t_{0})+\int_{t_{0}}^{\infty}(B_{D}+\delta)\delta dt=\infty,\end{equation}
 which contradicts the boundedness of $\eta(\theta)$. Since the claim
holds for all $\delta>0$, the result follows.

\bibliographystyle{abbrvnat}
\bibliography{ACFA}

\end{document}